\documentclass[lettersize,journal]{IEEEtran}

\usepackage{amsmath,amssymb,amsfonts}
\usepackage{textcomp}

\usepackage{graphicx}
\usepackage{float}
\usepackage{stfloats}
\usepackage{placeins}
\usepackage{adjustbox}
\usepackage{booktabs}
\usepackage{multirow}
\usepackage{makecell}
\usepackage{tabularx}
\usepackage{array}
\usepackage{siunitx}

\usepackage[table,xcdraw]{xcolor}

\usepackage{algorithm}
\usepackage{algpseudocode}

\definecolor{notegreen}{RGB}{165,195,188}
\algrenewcommand\algorithmiccomment[1]{%
  \hfill{\color{notegreen}$\triangleright$~#1}%
}

\usepackage{caption}
\usepackage{subcaption}
\usepackage{hyperref}
\usepackage{cite}
\usepackage{url}
\usepackage{cleveref}
\usepackage{tabularx}
\usepackage{array}

\newcolumntype{Y}{>{\centering\arraybackslash}X}

\usepackage{pifont}
\usepackage{bbding}
\usepackage{verbatim}
\usepackage{lipsum}
\usepackage{booktabs}
\usepackage[table]{xcolor}
\definecolor{oursrow}{HTML}{EAF7E6}
\newcommand{\cmark}{\ding{51}}  
\newcommand{\xmark}{\ding{55}}  

\begin{document}

\title{PhysFlow: Frequency Decoupled with Dual-Field Rectified Flow for Remote Photoplethysmography}

\author{Zixu Li, Jianjun Qian~\IEEEmembership{Member,~IEEE}, Hang Shao, Lei Luo, and Jian Yang
\thanks{Zixu Li, Jianjun Qian, Lei Luo and Jian Yang are with the PCA Lab, Key Lab of Intelligent Perception and Systems for High-Dimensional Information of Ministry of Education, School of Computer Science and Engineering, Nanjing University of Science and Technology, Nanjing 210094, China. Hang Shao is with the College of Computer Science and Technology, Qingdao University, Qingdao 266071, China.
(email:lizixu@njust.edu.cn; csjqian@njust.edu.cn; shaohang@qdu.edu.cn; cslluo@njust.edu.cn; csjyang@njust.edu.cn).

Copyright © 20xx IEEE. Personal use of this material is permitted. However, permission to use this material for any other purposes must be obtained from the IEEE by sending an email to pubs-permissions@ieee.org.
}

}

\markboth{Journal of \LaTeX\ Class Files,~Vol.~14, No.~8, August~2021}%
{Shell \MakeLowercase{\textit{et al.}}: A Sample Article Using IEEEtran.cls for IEEE Journals}

\IEEEpubid{}

\maketitle

\begin{abstract}

Remote Photoplethysmography (rPPG) enables contactless pulse estimation from facial videos, serving as a vital tool for health monitoring. However, current deep learning methods often struggle under complex disturbances, particularly varying illumination, facial expressions, and unconstrained head movements. In such scenarios, subtle physiological signals are easily dominated by external interference, making the recovered rPPG waveform unstable and unreliable. One important reason is that most existing methods directly model the rPPG signal in a unified manner, where different signal components are coupled during reconstruction. This makes it difficult to preserve weak pulse-related variations when strong disturbance-induced changes are present. To address this challenge, we propose PhysFlow, a frequency-decoupled dual-field rectified flow framework tailored for robust rPPG estimation. Specifically, the ground-truth rPPG signal is decomposed into trend and amplitude components, which are used as separate supervisory targets. Based on the extracted facial features, PhysFlow learns two component-specific conditional velocity fields to model the two components separately. This design reduces mutual interference between different components and improves the robustness of rPPG reconstruction under complex disturbances. Moreover, the rectified flow formulation enables efficient waveform reconstruction with only a few ordinary differential equation (ODE) integration steps. Extensive experiments on multiple benchmark datasets demonstrate that PhysFlow outperforms state-of-the-art methods in both heart-rate estimation and rPPG waveform reconstruction across diverse challenging scenarios.

\end{abstract}



\begin{figure}[h]
  \centering
  \includegraphics[width=0.985\linewidth]{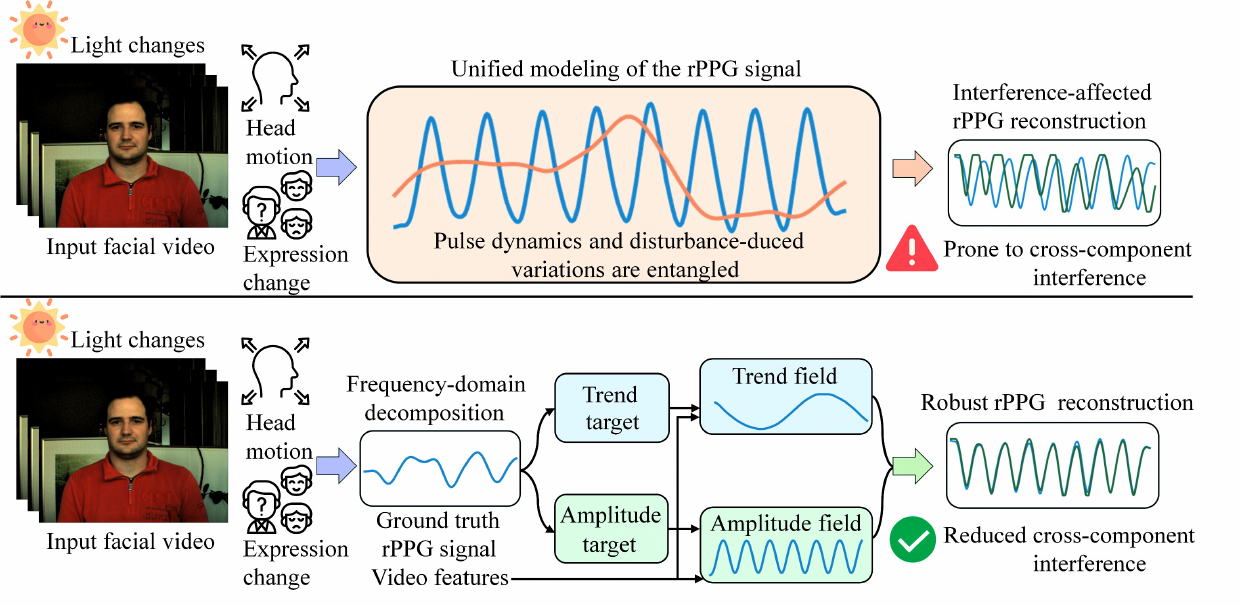}
  \caption{Comparison between conventional unified modeling and PhysFlow under complex disturbances. Unified modeling entangles pulse and disturbance variations, whereas PhysFlow performs separate trend and amplitude modeling to reduce cross-component interference and improve reconstruction robustness.}
     \label{fig:fig1}

\end{figure}
\section{Introduction}
\IEEEPARstart
Remote photoplethysmography (rPPG) is a contactless physiological sensing technique that estimates the blood volume pulse (BVP) by analyzing subtle, pulse-synchronous color and intensity variations in facial videos, thereby enabling the estimation of vital signs such as heart rate, respiratory rate and blood pressure \cite{1,2,55}. These fluctuations originate from heartbeat-driven changes in blood volume in superficial skin vessels, which induce subtle variations in the skin’s reflected-light intensity recorded by cameras \cite{3,4,54}. While invisible to the naked eye, the variations are measurable and can be processed to recover pulse-related physiological signals\cite{5}. Due to its non-invasive nature and ease of deployment, rPPG has been investigated and increasingly used in applications including driver monitoring, health surveillance and face anti-spoofing\cite{6,7,8,56}.

Early rPPG methods relied primarily on physiological and optical priors to recover pulse-related signals from facial videos~\cite{12,13}. These methods are usually built upon the observation that cardiac activity induces subtle periodic color variations on human skin, especially in regions with rich superficial blood perfusion. Based on this assumption, conventional approaches typically follow a hand-crafted signal processing pipeline, including facial region selection, skin color trace extraction, color space projection, spatial aggregation, temporal band-pass filtering, and source separation~\cite{9,11}. Representative methods exploit chrominance-based constraints, orthogonal skin-tone projection, or blind source separation to suppress illumination variations and isolate pulse-sensitive components from facial appearance changes. Owing to their clear physical motivation and low computational cost, these methods can achieve competitive performance under controlled acquisition conditions and standard benchmark settings~\cite{14,15,42,43}. However, their effectiveness is highly dependent on the validity of manually designed assumptions, such as stable illumination, limited head motion, reliable skin-region extraction, and relatively stationary pulse-related frequency components. In practical scenarios, these assumptions are often violated. Severe illumination changes introduce large-scale intensity fluctuations, free head movements alter facial reflectance and region correspondence, and rich facial expressions produce non-rigid appearance variations. These disturbance factors may overlap with the weak pulse-induced color changes in both temporal and spectral domains, making it difficult for hand-crafted filtering or projection strategies to separate physiological signals from nuisance components. As a result, the recovered rPPG waveform may suffer from spectral contamination, unstable morphology, and pronounced amplitude collapse~\cite{16}. To better capture the dynamic characteristics of rPPG signals, PhysDiff~\cite{17} introduces an amplitude--trend decoupling strategy based on temporal differencing between adjacent frames, which enhances the model's sensitivity to instantaneous signal variations. Nevertheless, robust reconstruction under complex disturbances remains challenging, especially when different signal components are still not explicitly modeled as separate reconstruction targets.

Although existing rPPG methods have achieved notable progress across diverse and challenging real-world scenarios, they still encounter pronounced challenges under complex disturbances, such as severe illumination variations, unconstrained head motion and rich facial expressions~\cite{18,19,20}. These challenges arise primarily from three aspects. First, most existing methods estimate rPPG signals directly from facial videos in an end-to-end manner, where weak yet physiologically meaningful pulse-related color variations are inevitably entangled with strong illumination and motion interference~\cite{50,52,53}, as illustrated in Fig.~\ref{fig:fig1}. When such interference overwhelms the subtle appearance variations induced by blood-volume dynamics, the learned representations are easily dominated by nuisance factors, thereby substantially compromising the accuracy and robustness of rPPG estimation. Second, even when trend-amplitude disentanglement~\cite{17} is introduced to enhance robustness, the decomposed signals are typically exploited only as auxiliary cues for denoising and reconstruction, rather than being explicitly formulated as distinct generation targets. Consequently, the potential of decomposition is not fully realized during the generative process. Moreover, diffusion-based inference still depends on iterative denoising and repeated sampling, resulting in considerable latency and computational overhead that hinder efficient and practical deployment. Third, waveform reconstruction quality remains insufficiently explored in rPPG estimation. Existing methods and evaluation protocols primarily emphasize HR-related accuracy, while the morphological fidelity of the reconstructed rPPG waveform is often underexamined. However, accurate HR estimation does not necessarily guarantee faithful waveform recovery. In challenging scenarios, the dominant cardiac rhythm may remain identifiable, yet the reconstructed signal can still suffer from oversmoothing or distortion in its overall shape, temporal structure and subtle pulsatile variations. As these waveform characteristics contain meaningful physiological and discriminative cues~\cite{45}, HR-related metrics alone are inadequate for a comprehensive and reliable evaluation of rPPG estimation quality.

To address these limitations, we propose PhysFlow, a dual-field rectified-flow framework for robust and efficient rPPG estimation under complex disturbances. Specifically, we decompose the ground-truth rPPG signal into a trend component and an amplitude component, which respectively capture the slowly varying waveform profile and fine-grained pulse-related variations. Rather than using these decomposed signals as auxiliary guidance, PhysFlow treats them as explicit supervisory targets during training and learns separate conditional velocity fields to generate the corresponding components from Gaussian-initialized latent states. In this manner, the two components are learned and reconstructed under distinct objectives, allowing the decomposition to participate explicitly in the generative process rather than serving only as an aid for reconstruction. This design not only enhances robustness against external disturbances, but also improves the preservation of the overall waveform shape and subtle pulse morphology during reconstruction. Furthermore, by adopting the velocity-matching formulation of rectified flow, PhysFlow avoids iterative denoising and directly learns the transport from Gaussian-initialized states to the target states corresponding to the trend and amplitude components. During inference, the final rPPG signal is reconstructed using only a few ordinary differential equation (ODE) integration steps, thereby further reducing computational overhead and improving the practicality of real-time deployment.

 Overall, our contributions are summarized as follows:
\begin{itemize} 
\item[$\bullet$] 
We propose PhysFlow, a frequency-decoupled dual-field rectified flow framework for robust rPPG estimation. Unlike conventional methods, PhysFlow separates the trend and amplitude components and models them via two distinct conditional velocity fields.

\item[$\bullet$] 
We propose a frequency-domain decomposition strategy to derive explicit supervisory targets for the trend and amplitude components, and introduce two component-specific velocity heads to model them separately, thereby reducing mutual interference.

\item[$\bullet$] 
Extensive experiments on benchmark datasets demonstrate that PhysFlow outperforms existing state-of-the-art methods across challenging scenarios.

\end{itemize}

\section{Related Work}
\label{sec:related_work}

\subsection{Deep Learning-based rPPG Methods}
Traditional rPPG approaches primarily rely on hand-crafted operations, including color-space transformations and blind source separation, to extract pulse signals with a high signal-to-noise ratio \cite{21,22}. However, their effectiveness degrades markedly in the presence of strong external disturbances. The difficulty stems from the intrinsically subtle skin-colour variations associated with blood volume dynamics. In recent years, deep learning has become the prevailing paradigm for rPPG estimation\cite{23,24,26,27}. Existing deep learning-based methods can be generally categorized into two types. The first one performs end-to-end regression of rPPG signals directly from facial videos. The second category constructs a Multi-scale Spatio-temporal map (MSTmap) using manually designed transformations and then learns a mapping from this representation to rPPG\cite{28,29,30,44}. Despite the notable progress achieved by these approaches in controlled settings, their robustness in complex and unconstrained scenarios remains limited. The PhysDiff \cite{17} introduces a diffusion-based rPPG estimation framework with physiology-inspired dynamic disentanglement to improve robustness against environmental perturbations. However, its generalization to challenging real-world scenarios remains limited. Recently, a transformer-based end-to-end rPPG predictor\cite{31} has been proposed to further improve performance under extreme conditions. Nevertheless, these advances still do not fully resolve the fundamental difficulty of robust rPPG estimation in complex real-world scenarios. In particular, maintaining stable signal reconstruction under complex disturbances while preserving efficient inference remains an open challenge.

\subsection{Rectified Flow}
Rectified Flow was first introduced in \cite{32}, where the transport process between two data distributions is straightened into an approximately linear trajectory, thereby enabling high-quality generation with only a few Euler integration steps. FlowGrad was subsequently proposed in \cite{33}, which further improves controllability by optimizing the ODE trajectory through a guidance function. More recently, Rectified Flow has been extended to low-level vision tasks in \cite{35,36}, where it improves inference efficiency while maintaining generation quality. For example, FlowIE was proposed in \cite{34}, which formulates a conditional Rectified Flow framework with a linear many-to-one transport mapping and achieves real-time image super-resolution using only five sampling steps. Despite these advances, the application of Rectified Flow to rPPG estimation remains largely underexplored. Motivated by this gap, we propose PhysFlow, a dual-field Rectified Flow model tailored for rPPG estimation.

\begin{figure*}[t]
  \centering
   \includegraphics[width=0.985\linewidth]{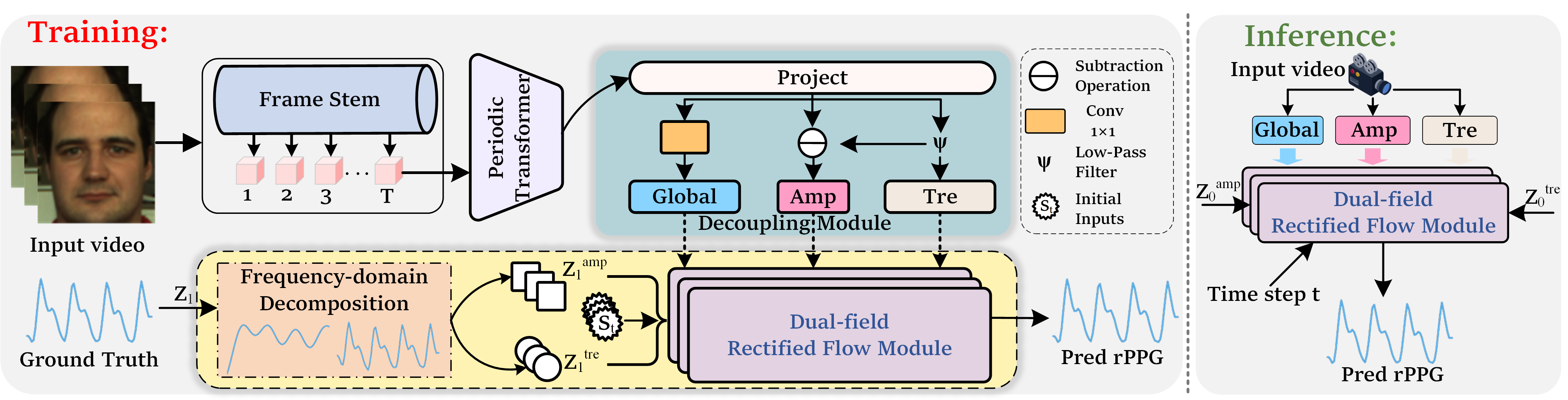}
   \caption{
Overview of the proposed PhysFlow. The ground-truth rPPG signal is decomposed into trend and amplitude components for separate supervision of two conditional velocity fields in a dual-field rectified-flow framework. The final rPPG signal is reconstructed from the generated trend and amplitude components. Here, the label Initial Inputs denotes the input tuple $(z_0^{\mathrm{amp}}, z_0^{\mathrm{tre}}, t)$, where $z_0^{\mathrm{amp}}$ and $z_0^{\mathrm{tre}}$ are the Gaussian-initialized latent states for the amplitude and trend components, and $t$ denotes the time step.}
   \label{Pic_Overview}
\end{figure*}

\section{Preliminary}
Traditional diffusion models are trained by injecting noise in the forward process, enabling the model to generate high-quality images from Gaussian noise. However, this procedure typically requires multiple sampling steps, which substantially increases inference time. In general, the forward process can be formulated as follows:
\begin{equation}
	x_t = a_t x_0 + b_t \gamma,\quad \gamma \sim \mathcal{N}(0,I),
\end{equation}
where $a_t$ and $b_t$ satisfy $a_t=0$ when $b_t=1$, and $b_t=0$ when $a_t=1$. In DDPM, this expression can be written as:
\begin{equation}
x_{t}=\sqrt{\bar{a_{t}} }x_{0} +\sqrt{1-\bar{a_{t}}} \gamma , \quad \gamma \sim \mathcal{N}(0,I).
\end{equation}
Unlike DDPM, which learns generation through progressive noise perturbation and typically requires many sampling steps, Rectified Flow (RF) learns a velocity field to transport samples from Gaussian distribution to the data distribution. The objective of  Rectified Flow is to optimize a model $v_{\theta}$ to approximate the velocity $v_{t}(x_t)$ on the path at step $t$. This is expressed by the following equation:
\begin{equation}
	\mathcal{L}_{\mathrm{RF}}(\theta)
	=
	\mathbb{E}_{x_0,x_1,t}\left\|v_{\theta}(x_t,t)-v_t(x_t)\right\|_2^2,
      \label{eq:3}
\end{equation}
for a coupled pair $(x_0,x_1)$ with $x_0\sim\pi_0$ and $x_1\sim\pi_1$, RF defines a straight-line path:
\begin{equation}
	x_t=(1-t)x_0+t x_1,\qquad t\in[0,1],
\end{equation}
under this construction, $v_{t}(x_t)$ can be written as:
\begin{equation}
	v_t(x_t)=\frac{d x_t}{dt}=x_1-x_0,
      \label{eq:5}
\end{equation}
then substituting Eq. \ref{eq:5} into the velocity-matching objective in Eq. \ref{eq:3}, the Rectified Flow training objective can be rewritten as:
\begin{equation}
	\mathcal{L}_{\mathrm{RF}}(\theta)
	=
	\mathbb{E}_{x_0,x_1,t}\left\|v_{\theta}(x_t,t)-(x_1-x_0)\right\|_2^2.
\end{equation}
By training the neural network on large-scale datasets, the network's output $v_{\theta}(x_t,t)$ is encouraged to closely approximate the training target $x_1-x_0$. This enables the model to find the shortest path between two data distributions, thereby accelerating the sampling process.

\section{Method}
\label{sec:method}

\subsection{PhysFlow Framework}

In this section, we provide a detailed description of the proposed PhysFlow framework for robust rPPG estimation. Unlike conventional methods that reconstruct the rPPG signal in a unified manner, PhysFlow explicitly decouples the supervision and modeling of different signal components to improve robustness under complex disturbances. As illustrated in Fig.~\ref{Pic_Overview}, the proposed framework mainly consists of four parts: (1) frequency-domain decomposition for constructing trend and amplitude supervisory targets; (2) periodic transformer block for extracting video guidance features; (3) dual-field rectified-flow module for learning component-specific conditional velocity fields; (4) training and inference procedure for reconstructing the final rPPG signal. The overall procedure is summarized in Algorithm~\ref{alg:physflow}.

\subsection{Problem Formulation}
In this work, we focus on leveraging the above Rectified Flow formulation to enable more efficient rPPG estimation. Let $z_1\in\mathbb{R}^{T}$ denote the ground-truth rPPG signal. During training, the rPPG signal is decomposed in the frequency domain into trend and amplitude components:
\begin{equation}
(z_1^{\mathrm{tre}},z_1^{\mathrm{amp}}) = FDD(z_1),
\end{equation}
where FDD$(\cdot)$ denotes a frequency-domain decomposition operator. Specifically, the input rPPG signal is first transformed into the frequency domain via the real Fast Fourier Transform (RFFT). A low-frequency mask and a band-pass mask are then applied to obtain the trend and amplitude components, which are finally mapped back to the time domain via the inverse real Fast Fourier Transform (IRFFT). The complete decomposition procedure is summarized in Algorithm~\ref{alg:fdd}.

The FDD$(\cdot)$ is not intended to achieve strict physiological source separation. Instead, it serves as a physiology-guided supervisory decomposition applied to the ground-truth rPPG signal. The rationale is that, despite subject and scenario variations, the trend component and the amplitude component usually exhibit distinct temporal characteristics: the former mainly captures the slowly varying waveform profile, whereas the latter mainly preserves pulse-related variations within the plausible heart-rate range. Therefore, the role of FDD$(\cdot)$ is not exact physiological factorization, but an interpretable and stable decomposition that provides two supervisory targets for the two conditional velocity fields. Unless otherwise specified, we set $f_{tr}=0.5$ Hz, $f_{am}=0.7$ Hz, $f_{\max}=3.0$ Hz, and $\tau=0.10$ Hz, where $\tau$ controls the smoothness of the spectral transition.

We initialize the trend and amplitude latent states with Gaussian noise. Given the target states $z_1^{\mathrm{tre}}$ and $z_1^{\mathrm{amp}}$, we construct linear interpolation paths from the initial noise states to the corresponding targets for the two components. This process can be formulated as:
\begin{equation}
	\begin{gathered}
	z_{t}^{\mathrm{tre}}=(1-t)z_{0}^{\mathrm{tre}}+t\,z_1^{\mathrm{tre}},\quad z_{0}^{\mathrm{tre}}\sim\mathcal{N}(0,I),\\
	z_{t}^{\mathrm{amp}}=(1-t)z_{0}^{\mathrm{amp}}+t\,z_1^{\mathrm{amp}},\quad 	z_{0}^{\mathrm{amp}}\sim\mathcal{N}(0,I), 
	\end{gathered}
    \label{eq:8}
\end{equation}
differentiating the interpolation paths in Eq. \ref{eq:8} with respect to $t$ yields the corresponding target velocities in the trend and amplitude components:
\begin{equation}
	\begin{gathered}
		v_t^{tre}(x_t)=\frac{d (z_{t}^{\mathrm{tre}})}{d t} = z_1^{\mathrm{tre}} - z_{0}^{\mathrm{tre}},\\
		v_t^{amp}(x_t)=\frac{d (z_{t}^{\mathrm{amp}})}{d t} = z_1^{\mathrm{amp}} - z_{0}^{\mathrm{amp}},\\
	\end{gathered}
    \label{eq:9}
\end{equation}
where $t\in[0,1]$ denotes the continuous interpolation time. To model the corresponding transport process, we further define two conditional velocity fields for the trend and amplitude components, denoted as $v_{\theta}^{\mathrm{tre}}$ and $v_{\theta}^{\mathrm{amp}}$:
\begin{equation}
	\begin{gathered}
v_{\theta}^{\mathrm{amp}}=f_{\theta}^{\mathrm{amp}}\!\big(z_t^{\mathrm{amp}},t,h_{\mathrm{glb}},c_{\mathrm{amp}}),\\
v_{\theta}^{\mathrm{tre}}=f_{\theta}^{\mathrm{tre}}\!\big(z_t^{\mathrm{tre}},t,h_{\mathrm{glb}},c_{\mathrm{tre}}).\\
	\end{gathered}
\end{equation}
Where $f_{\theta}^{\mathrm{amp}}(\cdot)$ and $f_{\theta}^{\mathrm{tre}}(\cdot)$ denote the proposed Amplitude head and Trend head respectively, which predict the conditional velocities in the amplitude and trend components based on the current state, interpolation time and guidance features. The guidance features $h_{\mathrm{glb}}$, $c_{\mathrm{amp}}$ and $c_{\mathrm{tre}}$ are extracted from the input video V $\in \mathbb{R}^{T \times 3 \times H \times W}$, where T denotes the number of frames, $H$ and $W$ represent the spatial resolution. Specifically, $h_{\mathrm{glb}}$ denotes the global feature shared by both branches, while $c_{\mathrm{amp}}$ and $c_{\mathrm{tre}}$ denote the amplitude-specific and trend-specific guidance features.

After learning the two conditional velocity fields, PhysFlow reconstructs the final rPPG signal at inference time by separately evolving the amplitude and trend latent states. Starting from the initial states $z_0^{\mathrm{amp}}$ and $z_0^{\mathrm{tre}}$, the two variables are updated from $t=0$ to $t=1$ using $S$ uniformly discretized ODE steps with step size $\Delta t = 1/S$ and discrete time points $t_i=i\Delta t$. At each step, the conditional velocities are evaluated based on the current latent states and then integrated using the explicit Euler scheme:

\begin{equation}
	\begin{gathered}
		z_{i+1}^{\mathrm{amp}} = z_i^{\mathrm{amp}} + \Delta t \cdot f_{\theta}^{\mathrm{amp}}\!\big(z_i^{\mathrm{amp}}, t_i, h_{\mathrm{glb}}, c_{\mathrm{amp}}\big),\\
		z_{i+1}^{\mathrm{tre}} = z_i^{\mathrm{tre}} + \Delta t \cdot f_{\theta}^{\mathrm{tre}}\!\big(z_i^{\mathrm{tre}}, t_i, h_{\mathrm{glb}}, c_{\mathrm{tre}}\big),
	\end{gathered}
	\label{eq:sampling}
\end{equation}
the updates are performed for $i=0,1,\dots,S-1$. Repeating this process for $S$ steps yields the terminal latent states $z_S^{\mathrm{amp}}$ and $z_S^{\mathrm{tre}}$. These two terminal states are finally passed to the Predictor head to reconstruct the output rPPG signal:
\begin{equation}
	z_{\mathrm{out}}=\mathrm{Predictor}\!\left(z_S^{\mathrm{amp}}, z_S^{\mathrm{tre}}\right).
	\label{eq:decoder}
\end{equation}

\begin{algorithm}[t]
\caption{Overall algorithm of PhysFlow}
\label{alg:physflow}
\footnotesize
\begin{algorithmic}[1]
\State \textbf{Input:} facial video $V$ and ground truth signal $z_1$
\State \textbf{Output:} reconstructed signal $z_{\mathrm{out}}$

\State $\{h_{\mathrm{glb}},c_{\mathrm{amp}},c_{\mathrm{tre}}\}\gets \Phi(V)$
    \Comment{video guidance features}

\State \textbf{Training phase:}
\State $\{z_1^{\mathrm{amp}},z_1^{\mathrm{tre}}\}\gets \mathrm{FDD}(z_1)$, 
       $\{z_0^{\mathrm{amp}},z_0^{\mathrm{tre}}\}\sim\mathcal{N}(0,I)$, 
       $t\sim\mathcal{U}(0,1)$
\State Construct $z_t^{\mathrm{amp}}$ and $z_t^{\mathrm{tre}}$, then predict $\hat{v}_{\theta}^{\mathrm{amp}}$ and $\hat{v}_{\theta}^{\mathrm{tre}}$
\State \textbf{Loss:} $\mathcal{L}_{\mathrm{total}}
= \mathcal{L}_{\mathrm{RF}}(\theta)
+\lambda_{\mathrm{np}}\mathcal{L}_{\mathrm{NP}}$
    \Comment{loss optimization}

\State \textbf{Inference phase:}
\State Initialize $\{z^{amp}_0,z^{tre}_0\}\sim \mathcal{N}(0,I)$ with a fixed evaluation seed
\For{$s=0$ to $S-1$}
\State Predict $\hat{v}_{\theta}^{\mathrm{amp}}$ and $\hat{v}_{\theta}^{\mathrm{tre}}$, then update $z_s^{\mathrm{amp}}$ and $z_s^{\mathrm{tre}}$ via ODE integration steps
\EndFor
\State $z_{\mathrm{out}}\gets \mathrm{Predictor}(z_S^{\mathrm{amp}},z_S^{\mathrm{tre}})$
    \Comment{signal reconstruction}
\end{algorithmic}
\end{algorithm}

\begin{figure}[t]   
  \centering
  \includegraphics[width=0.985\linewidth]{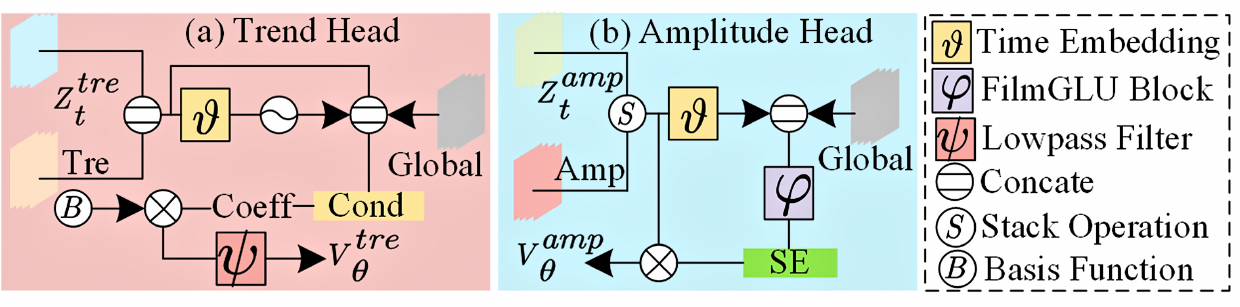} 
\caption{Illustration of the proposed velocity heads in PhysFlow. (a) shows the Trend Head for modeling the trend component. (b) shows the Amplitude Head for modeling the amplitude component.}
  \label{fig:fig3}
\end{figure}
\subsection{Periodic Transformer Block}

After the frame stem, the input video is converted into a temporal token sequence. Since rPPG estimation fundamentally depends on modeling subtle periodic physiological variations, multi-scale temporal representations are particularly suitable for capturing such pulsatile dynamics. Following prior work \cite{Rhythmformer}, we employ a multi-scale temporal periodic module as the feature extractor to generate guidance features. Through its hierarchical stage-wise processing, this module progressively refines temporal dependencies and enhances periodic cues that are critical for accurate rPPG estimation. In addition, we introduce an explicit decoupling module to separate global, amplitude-specific and trend-specific features, which are used to guide the learning of the trend and amplitude velocity heads.

\begin{algorithm}[t]
\caption{Frequency-domain Decomposition (FDD)}
\label{alg:fdd}
\footnotesize
\begin{algorithmic}[1]
\Require Ground-truth rPPG signal $z_1\in\mathbb{R}^{T}$, sampling rate $f_s$, cutoffs $f_{\mathrm{tre}}, f_{\mathrm{amp}}, f_{\max}$, softness $\tau$
\Ensure Trend target $z_1^{\mathrm{tre}}$, amplitude target $z_1^{\mathrm{amp}}$

\State $Z \gets \mathrm{RFFT}(z_1), \quad \mathbf{f} \gets \mathrm{RFFTFreq}(T,f_s)$

\State $M^{\mathrm{tre}} \gets \sigma\!\left((f_{\mathrm{tre}}-\mathbf{f})/\tau\right)$
\Comment{trend mask}

\State $M^{\mathrm{amp}} \gets \sigma\!\left((\mathbf{f}-f_{\mathrm{amp}})/\tau\right)\odot
\sigma\!\left((f_{\max}-\mathbf{f})/\tau\right)$
\Comment{amplitude mask}

\State $z_1^{\mathrm{tre}} \gets \mathrm{IRFFT}(Z\odot M^{\mathrm{tre}})$
\Comment{trend supervision}

\State $z_1^{\mathrm{amp}} \gets \mathrm{IRFFT}(Z\odot M^{\mathrm{amp}})$
\Comment{amplitude supervision}

\State \Return $z_1^{\mathrm{tre}}, z_1^{\mathrm{amp}}$
\end{algorithmic}
\end{algorithm}

\subsection{Dual-field Rectified Flow Module}
PhysFlow simultaneously evolves the amplitude latent state $z_t^{\mathrm{amp}}$ and the trend latent state $z_t^{\mathrm{tre}}$ in two parallel transport processes under the guidance of the extracted video features $h_{\mathrm{glb}}$, $c_{\mathrm{amp}}$ and $c_{\mathrm{tre}}$. At each interpolation time $t\in[0,1]$, the corresponding conditional velocities are predicted by the Amplitude Head and the Trend Head. The Amplitude Head models the amplitude-related variations under the guidance of $c_{\mathrm{amp}}$ and $h_{\mathrm{glb}}$, while the Trend Head models the trend component under the guidance of $c_{\mathrm{tre}}$ and $h_{\mathrm{glb}}$. The shared global feature $h_{\mathrm{glb}}$ provides additional global guidance for both branches.

Although both heads are formulated as conditional velocity predictors, they are not designed symmetrically because the two supervisory targets exhibit different temporal characteristics. The trend component mainly captures the global waveform profile and is dominated by low-frequency variations. Therefore, its velocity predictor should explicitly favor smoothness and suppress unnecessary high-frequency responses. In contrast, the amplitude component mainly preserves pulse-related variations and is more sensitive to video-dependent disturbances. Its velocity predictor therefore requires stronger condition awareness and temporal modeling flexibility. Accordingly, we introduce different architectural biases into the two heads rather than using an identical predictor for both components. In the following, we describe the two velocity heads.

\textbf{Trend Head:}
As illustrated in Fig.~\ref{fig:fig3} (a), the Trend Head is designed to predict the velocity of the trend component. Since this component mainly captures the global waveform profile, directly using a highly flexible temporal predictor may introduce unnecessary high-frequency fluctuations. To impose a smoothness prior, we regress a compact set of basis coefficients and reconstruct the trend velocity in a low-dimensional temporal subspace.

Given the current trend latent state $z_t^{\mathrm{tre}}$ and the trend condition $c_{\mathrm{tre}}$, we first concatenate them to form a fused representation:
\begin{equation}
	z_c=\mathrm{Concat}(z_t^{\mathrm{tre}},\,c_{\mathrm{tre}}),
	\label{eq:trend_concat}
\end{equation}
where the continuous interpolation time $t$ is encoded by sinusoidal time embedding $\phi(t)$. The basis coefficients are then estimated from the fused trend representation, the global video feature, and the time embedding, which can be formulated as:
\begin{equation}
	\begin{gathered}
	g=\mathrm{GELU}\big(\mathrm{Linear}(z_c)\big),\\
	\mathrm{Coeff}=\mathrm{MLP}\big(\mathrm{Concat}(g,h_{\mathrm{glb}},\phi(t),\,z_c)\big),\\
	\tilde v_{\theta}^{\mathrm{tre}}=\mathrm{Coeff} \otimes B,
	\end{gathered}
\end{equation}
where B is a hybrid basis matrix constructed by concatenating spectral bases and polynomial bases:
\begin{equation}
	B = [B_{\mathrm{spe}}, B_{\mathrm{poly}}].
\end{equation}
The spectral bases provide smooth periodic and low-frequency patterns, while the polynomial bases capture slowly varying baseline drifts. After basis reconstruction, a low-pass operator LP$(\cdot)$ is further applied to suppress residual high-frequency components. The final trend velocity is obtained as:
\begin{equation}
	v_{\theta}^{\mathrm{tre}} = \lambda \cdot \tanh\!\Big(\beta \cdot LP(\tilde v_{\theta}^{\mathrm{tre}})\Big).
\end{equation}
Here, $\lambda$ bounds the maximum velocity magnitude and $\beta$ controls the saturation strength of the $\tanh$ activation, which stabilizes the transport process. In all experiments, we fix $\lambda=3.0$ and $\beta=0.33$.

\textbf{Amplitude Head:}
As shown in Fig.~\ref{fig:fig3} (b), the Amplitude Head is designed to predict the velocity of the amplitude component. Unlike the slowly varying trend component, the amplitude component mainly preserves pulse-related variations and is more sensitive to video-dependent disturbances. Accordingly, we employ a condition-aware temporal predictor, in which FiLM modulation incorporates the interpolation time and global video context, dilated temporal convolutions enlarge the temporal receptive field, and SE reweighting adaptively emphasizes informative temporal responses.

Specifically, we first construct a two-channel temporal input by stacking the current amplitude latent state and the amplitude condition, and then project it into a hidden space:
\begin{equation}
	x_0=\mathrm{Linear}\big(\mathrm{Stack}[z_t^{\mathrm{amp}},\,c_{\mathrm{amp}}]\big),
	\label{eq:amp_proj}
\end{equation}
where a FiLM modulation vector is computed from the continuous time embedding and the global video feature:
\begin{equation}
	\gamma(t)=\tanh\!\big(\mathrm{Linear}(\mathrm{Concat}(\phi(t),\,h_{\mathrm{glb}}))\big),
	\label{eq:amp_gamma}
\end{equation}
the modulation vector allows the amplitude velocity predictor to adapt its temporal response along the rectified-flow trajectory under video-conditioned guidance. The projected sequence is then processed by $L_{th}$ GLUB temporal blocks with dilated depthwise convolutions:
\begin{equation}
	x_{\ell}=\mathrm{GLUB}\!\left(x_{\ell-1},\,\gamma(t)\right), \quad \ell=1,2,\dots,L_{th}.
\end{equation}
Here, the dilated convolutions enlarge the temporal receptive field, enabling the block to model pulse-related temporal dependencies over a broader range. The FiLMGLU block is defined as:
\begin{equation}
	\begin{gathered}
		x_{\beta}=\mathrm{Conv}_{3}\big(\mathrm{Norm}(\mathrm{Conv}_{1}(x_{\ell-1}))\big),\\
		(x_{1},x_{2})=\mathrm{split}(x_{\beta},\mathrm{dim}=1),\\
		x_{\ell}=x_{\ell-1}+x_{1}\ast \mathrm{Sigmoid}\!\big(x_{2}+\gamma(t)\big).
	\end{gathered}
\end{equation}
Finally, we apply squeeze-and-excitation (SE) reweighting to adaptively recalibrate the temporal features and output the amplitude velocity:
\begin{equation}
	\begin{gathered}
	s=\mathrm{Sigmoid}\!\left(SE(\mathrm{Mean}(x_{L_{th}}))\right),\\
	\hat{x}=s\odot x_{L_{th}},\\
	v_{\theta}^{\mathrm{amp}}
	=\delta \cdot \tanh\!\Big(\alpha\cdot \mathrm{Conv}_{1\times1}\big(\hat{x}\big)\Big).
	\end{gathered}
\end{equation}
Where $\mathrm{Mean}(\cdot)$ denotes temporal averaging. Similar to the Trend Head, $\alpha$ and $\delta$ are fixed scaling factors used to bound the amplitude velocity and stabilize the ODE integration. We set $\alpha=0.55$ and $\delta=8$ for all experiments.

\subsection{Loss Function}

To train the model, we employ the proposed dual-field Rectified Flow loss as the primary supervision and the NegPearson loss as an auxiliary objective. The overall training loss is defined as:
\begin{equation}
\mathcal{L}_{\mathrm{total}}
=
\mathcal{L}_{\mathrm{RF}}(\theta)
+
\lambda_{\mathrm{np}}\mathcal{L}_{\mathrm{NegPearson}}(z_{\mathrm{out}}, z_1),
\label{eq:overall_loss}
\end{equation}
where $\lambda_{\mathrm{np}}=0.25$, $z_{\mathrm{out}}\in\mathbb{R}^{T}$ denotes the predicted rPPG and $z_1\in\mathbb{R}^{T}$ denotes the ground-truth rPPG. We use Rectified Flow-based velocity matching for both trend and amplitude components by aligning each predicted velocity with its corresponding target velocity along the interpolation path. The dual-field Rectified Flow loss is formulated as:
\begin{equation}
	\begin{gathered}
\mathcal{L}_{\mathrm{RF}}(\theta)=
\mathbb{E}_{t,\{z_0^k,z_1^k\}}
\Big[
\big\|v_{\theta}^{\mathrm{tre}}(z_t^{\mathrm{tre}},t, h_{\mathrm{glb}},c_{\mathrm{tre}})
-(z_1^{\mathrm{tre}}-z_{0}^{\mathrm{tre}})\big\|_{2}^{2}
\\
+\big\|v_{\theta}^{\mathrm{amp}}(z_t^{\mathrm{amp}},t, h_{\mathrm{glb}},c_{\mathrm{amp}})
-(z_1^{\mathrm{amp}}-z_{0}^{\mathrm{amp}})\big\|_{2}^{2}
\Big].
	\end{gathered}
\end{equation}
Where $z_0^k$ is sampled from the standard Gaussian distribution and $z_1^k$ is obtained by decomposing the ground-truth rPPG signal into the corresponding component.

\section{Experiment}
\label{sec:exp}
In this section, we evaluate the proposed PhysFlow algorithm on four public benchmark datasets(namely BUAA-MIHR~\cite{39}, VIPL-HR~\cite{37,38}, NIRP-DRV~\cite{40-drv} and NIRP-IND~\cite{40-indoor}), and compare its performance with existing state-of-the-art methods. Both qualitative and quantitative experiments are conducted for rPPG-based heart rate estimation under intra-dataset and cross-dataset settings. In addition, we discuss the computational cost and parameter complexity of the proposed method, while ablation studies are performed to verify the effectiveness of its key modules. Finally, we summarize the conclusions of this paper and discuss its limitations.


\subsection{Datasets}
\textbf{BUAA-MIHR:} This dataset is a benchmark for rPPG evaluation under low-illumination conditions. It contains 165 RGB facial videos from 15 subjects, recorded under 11 illumination levels ranging from 1 to 100 lux. We only use samples with illumination intensities of at least 6.3 lux. Each video lasts 60 seconds at 30 fps with a resolution of 640$\times$480, and synchronized finger-clipped PPG signals sampled at 60 Hz are provided as ground truth.

\textbf{VIPL-HR:} This dataset contains 3,130 facial videos from 107 subjects, covering diverse head motions, illumination conditions, and acquisition devices. The recordings include both visible-light and near-infrared videos, and each video is approximately 30 seconds long. Synchronized finger PPG signals sampled at 60 Hz are used as ground truth.

\textbf{NIRP-DRV:} This dataset is designed for rPPG measurement in driving scenarios. It contains RGB and near-infrared facial videos from 18 subjects, recorded at 30 fps under diverse in-vehicle lighting conditions and natural head movements. Synchronized heart-rate ground truth is provided for evaluation.

\textbf{NIRP-IND:} This dataset provides a complementary indoor evaluation setting to NIRP-DRV. It contains RGB and narrow-band near-infrared facial videos from 9 subjects, recorded at 30 fps under controlled and varying indoor illumination conditions. Synchronized pulse oximeter readings are provided as heart-rate ground truth.

\subsection{Performance Metrics}
Following the mainstream evaluation protocol\cite{46}, we adopt three commonly used quantitative metrics for heart rate estimation: Mean Absolute Error (MAE), Root Mean Square Error (RMSE), and Pearson’s Correlation Coefficient ($\rho$). Specifically, MAE and RMSE are used to measure the accuracy of the estimated heart rate, and $\rho$ indicates the linear correlation between the predicted and ground-truth heart rates.

Besides HR estimation metrics, we also report waveform-level complementary metrics for reconstruction evaluation, including waveform correlation (Wave-Corr) for morphology, phase error (PE) for phase alignment, and signal-to-noise ratio (SNR) for spectral quality.

\textbf{Waveform Correlation (Wave-Corr):}
\begin{equation}
\mathrm{Wave\mbox{-}Corr}
=
\frac{1}{M}\sum_{m=1}^{M}\mathrm{Corr}(z_{gt}^{m},z_{pred}^{m}),
\end{equation}
where $z_{gt}^{m}$ and $z_{pred}^{m}$ denote the ground-truth and predicted rPPG waveforms of the $m$-th evaluated sequence. A higher value indicates better waveform morphology consistency.

\textbf{Phase Error (PE):}
\begin{equation}
\begin{aligned}
	\mathrm{PE}
	&=
	\frac{1}{M}\sum_{m=1}^{M}
	\frac{|\ell_m^{*}|}{f_s}, \\
	\ell_m^{*}
	&=
	\arg\max_{\ell}
	\mathrm{Corr}\bigl(
	z_{pred}^{m}(t+\ell), z_{gt}^{m}(t)
	\bigr).
\end{aligned}
\end{equation}
Where $\ell_m^{*}$ is the optimal temporal lag obtained by cross-correlation, and $f_s$ is the sampling rate. A lower value indicates better phase alignment.

\textbf{Signal-to-Noise Ratio (SNR):}
\begin{equation}
\mathrm{SNR}
=
\frac{1}{M}\sum_{m=1}^{M}
10\log_{10}
\frac{
\sum_{f\in\Omega_{HR}^{m}}P_{pred}^{m}(f)
}{
\sum_{f\in\Omega_{all}\setminus\Omega_{HR}^{m}}P_{pred}^{m}(f)
}.
\end{equation}
Where $P_{pred}^{m}(f)$ is the power spectrum of the predicted waveform, $\Omega_{HR}^{m}$ is the frequency band around the ground-truth HR frequency, and $\Omega_{all}$ denotes the valid physiological frequency range. A higher value indicates better spectral quality.

\begin{table*}[t]
  \caption{Intra-dataset comparison results (bpm) of remote HR estimation on NIRP-IND, NIRP-DRV, VIPL-HR, and BUAA-MIHR datasets. $\downarrow$ indicates that lower values are better, while $\uparrow$ indicates that higher values are better. The best results are highlighted in \textbf{bold}, and the second-best results are underlined.}
  \label{tab:intra_dataset}
  \centering
  \footnotesize
  \renewcommand{\arraystretch}{1.08}
  \setlength{\tabcolsep}{2.2pt}

  \begin{tabularx}{\textwidth}{@{}>{\raggedright\arraybackslash}p{0.22\textwidth} YYY YYY YYY YYY@{}}
    \toprule
    \multirow{2}{*}{\makecell{Methods / Venues}} &
    \multicolumn{3}{c}{NIRP-IND} &
    \multicolumn{3}{c}{NIRP-DRV} &
    \multicolumn{3}{c}{VIPL-HR} &
    \multicolumn{3}{c}{BUAA-MIHR} \\
    \cmidrule(lr){2-4} \cmidrule(lr){5-7} \cmidrule(lr){8-10} \cmidrule(lr){11-13}
    &
    MAE$\downarrow$ & RMSE$\downarrow$ & $\rho \uparrow$ &
    MAE$\downarrow$ & RMSE$\downarrow$ & $\rho \uparrow$ &
    MAE$\downarrow$ & RMSE$\downarrow$ & $\rho \uparrow$ &
    MAE$\downarrow$ & RMSE$\downarrow$ & $\rho \uparrow$ \\
    \midrule

    POS \cite{pos} / \textit{TBME'2017}
    & 5.52 & 6.85 & 0.40
    & 12.75 & 15.36 & 0.34
    & 11.50 & 17.20 & 0.30
    & 5.04 & 7.12 & 0.63 \\

    CHROM \cite{chrom} / \textit{TBME'2013}
    & 6.84 & 8.41 & 0.32
    & 14.52 & 17.41 & 0.18
    & 11.07 & 17.99 & 0.27
    & 6.09 & 8.29 & 0.51 \\

    LGI \cite{LGI} / \textit{CVPR'2018}
    & 8.95 & 11.01 & 0.39
    & 12.90 & 15.51 & 0.31
    & 12.84 & 19.02 & 0.29
    & 6.97 & 11.33 & 0.42 \\

    \addlinespace[2pt]

    DeepPhys \cite{DeepPhys} / \textit{ECCV'2018}
    & 3.11 & 4.44 & 0.74
    & 13.22 & 18.39 & 0.43
    & 11.00 & 13.80 & 0.11
    & 4.78 & 6.74 & 0.69 \\

    TS-CAN \cite{TS-CAN} / \textit{NeurIPS'2020}
    & 2.49 & 3.89 & 0.75
    & 12.70 & 18.03 & 0.47
    & 9.39 & 14.59 & 0.21
    & 4.84 & 6.89 & 0.68 \\

    PFE-TFA \cite{PFE-TFA} / \textit{AAAI'2023}
    & 2.81 & 4.57 & 0.84
    & 5.34 & 8.92 & 0.73
    & 6.91 & 8.65 & 0.65
    & 1.29 & 2.65 & 0.91 \\

    NEST \cite{NEST} / \textit{CVPR'2023}
    & 1.08 & 2.26 & 0.89
    & 3.61 & 7.32 & 0.82
    & 5.52 & 7.96 & 0.80
    & 2.88 & 4.69 & 0.89 \\

    ND-DeeprPPG \cite{16} / \textit{TIP'2024}
    & 0.66 & 1.45 & 0.89
    & 3.47 & 6.54 & 0.85
    & 5.15 & 7.52 & 0.78
    & \underline{0.58} & 1.81 & \underline{0.95} \\

    EfficientPhys \cite{Efficientphys} / \textit{WACV'2023}
    & 1.37 & 4.81 & 0.81
    & 3.67 & 12.28 & 0.81
    & 5.23 & 8.25 & 0.72
    & 1.43 & 4.98 & 0.93 \\

    PhysFormer++ \cite{Physformer++} / \textit{IJCV'2023}
    & 0.64 & 1.39 & 0.89
    & 3.56 & 7.59 & 0.83
    & 4.88 & 7.62 & 0.80
    & 0.93 & 1.66 & 0.91 \\

    RhythmFormer \cite{Rhythmformer} / \textit{PR'2025}
    & 0.52 & 1.10 & \underline{0.91}
    & 3.44 & 6.72 & 0.82
    & 4.75 & 7.49 & 0.82
    & 0.67 & 1.57 & 0.94 \\

    RhythmMamba \cite{Rhythmmamba} / \textit{AAAI'2025}
    & 0.61 & 1.24 & 0.90
    & 3.35 & 6.53 & 0.83
    & 4.27 & 7.12 & 0.82
    & 0.96 & 1.82 & 0.90 \\

    PhysDiff \cite{17} / \textit{AAAI'2025}
    & 0.50 & 1.06 & \underline{0.91}
    & 3.24 & 6.26 & 0.83
    & 3.92 & 6.65 & 0.85
    & 0.82 & 1.71 & 0.93 \\

    PHASE-Net \cite{43} / \textit{CVPR'2026}
    & \underline{0.48} & \underline{0.98} & \underline{0.91}
    & \underline{3.18} & \underline{6.23} & \underline{0.86}
    & \underline{3.87} & \underline{6.37} & \underline{0.86}
    & 0.63 & \underline{1.43} & 0.94 \\

    \rowcolor{oursrow}
    \textbf{PhysFlow (Ours)}
    & \textbf{0.43} & \textbf{0.89} & \textbf{0.92}
    & \textbf{2.63} & \textbf{5.93} & \textbf{0.87}
    & \textbf{3.64} & \textbf{5.79} & \textbf{0.87}
    & \textbf{0.46} & \textbf{1.17} & \textbf{0.96} \\

    \bottomrule
  \end{tabularx}
\end{table*}

\subsection{Implementation Details}
All experiments are conducted on RGB-only facial videos within the open-source rPPG-Toolbox framework~\cite{rppg-toolbox}. All data are processed using a unified pipeline, where the facial region is detected, cropped, resized to $128\times128$, and divided into clips of length $T$. The ground-truth physiological signal is temporally aligned with the video frames and resampled to the video frame rate. The model is trained for 30 epochs using AdamW with an initial learning rate of $3\times10^{-4}$ and a weight decay of $1\times10^{-5}$. During training, the ground-truth rPPG signal is decomposed into trend and amplitude components as separate supervisory targets. During inference, PhysFlow reconstructs the rPPG waveform from Gaussian-initialized latent states using fixed-step Euler ODE integration. Unless otherwise specified, all main results are reported with a single random Gaussian initialization under a fixed evaluation seed, without repeated sampling or best-of-(N) selection. For dataset partitioning, 3/4 of the subjects are used for training and the remaining 1/4 for testing. NIRP-IND and NIRP-DRV are jointly used under the NIRP setting, while VIPL-HR and BUAA are trained independently. Cross-dataset evaluation is conducted only between NIRP and VIPL-HR. HR is estimated from the dominant frequency of the predicted rPPG waveform within the predefined physiological frequency range.

\begin{table*}[t]
  \caption{Cross-dataset comparison results (bpm) of remote HR estimation. The models are trained on one dataset and tested on another dataset. $\downarrow$ indicates that lower values are better, while $\uparrow$ indicates that higher values are better. The best results are highlighted in \textbf{bold}, and the second-best results are underlined.}
  \label{tab:cross_dataset}
  \centering
  \footnotesize
  \renewcommand{\arraystretch}{1.08}
  \setlength{\tabcolsep}{2.2pt}

  \begin{tabularx}{\textwidth}{@{}>{\raggedright\arraybackslash}p{0.25\textwidth} YYY YYY YYY@{}}
    \toprule
    \multirow{2}{*}{\makecell{Methods / Venues}} &
    \multicolumn{3}{c}{NIRP $\rightarrow$ VIPL-HR} &
    \multicolumn{3}{c}{VIPL-HR $\rightarrow$ NIRP-IND} &
    \multicolumn{3}{c}{VIPL-HR $\rightarrow$ NIRP-DRV} \\
    \cmidrule(lr){2-4} \cmidrule(lr){5-7} \cmidrule(lr){8-10}
    &
    MAE$\downarrow$ & RMSE$\downarrow$ & $\rho \uparrow$ &
    MAE$\downarrow$ & RMSE$\downarrow$ & $\rho \uparrow$ &
    MAE$\downarrow$ & RMSE$\downarrow$ & $\rho \uparrow$ \\
    \midrule

    DeepPhys \cite{DeepPhys} / \textit{ECCV'2018}
    & 15.53 & 17.48 & 0.41
    & 6.58 & 9.16 & 0.52
    & 10.51 & 12.71 & 0.56 \\

    TS-CAN \cite{TS-CAN} / \textit{NeurIPS'2020}
    & 12.04 & 15.12 & 0.45
    & 6.57 & 9.25 & 0.57
    & 10.37 & 12.65 & 0.61 \\

    PFE-TFA \cite{PFE-TFA} / \textit{AAAI'2023}
    & 8.20 & 11.22 & 0.64
    & 1.87 & 3.67 & 0.77
    & 7.82 & 9.99 & 0.64 \\

    NEST \cite{NEST} / \textit{CVPR'2023}
    & 5.15 & 8.78 & 0.72
    & 2.78 & 4.63 & 0.78
    & 6.27 & 8.45 & 0.65 \\

    ND-DeeprPPG \cite{16} / \textit{TIP'2024}
    & 5.08 & 7.92 & \underline{0.75}
    & 1.74 & 3.28 & 0.81
    & 6.18 & 7.75 & 0.68 \\

    EfficientPhys \cite{Efficientphys} / \textit{WACV'2023}
    & 5.25 & 11.81 & 0.71
    & 1.80 & 6.19 & 0.76
    & 6.07 & 11.53 & 0.64 \\

    PhysFormer++ \cite{Physformer++} / \textit{IJCV'2023}
    & 5.09 & 8.97 & 0.71
    & 1.79 & 3.77 & 0.81
    & 5.78 & 8.65 & 0.66 \\

    RhythmFormer \cite{Rhythmformer} / \textit{PR'2025}
    & 4.83 & 8.05 & 0.74
    & 1.70 & 3.43 & 0.83
    & 5.40 & 7.93 & 0.71 \\

    RhythmMamba \cite{Rhythmmamba} / \textit{AAAI'2025}
    & 4.30 & \underline{7.49} & \underline{0.75}
    & 1.76 & 3.51 & 0.84
    & \underline{5.31} & \underline{7.52} & \underline{0.72} \\

    PhysDiff \cite{17} / \textit{AAAI'2025}
    & 4.34 & 8.12 & 0.74
    & 1.62 & 3.35 & 0.85
    & 5.46 & 7.96 & \underline{0.72} \\

    PHASE-Net \cite{43} / \textit{CVPR'2026}
    & \underline{4.27} & 7.96 & \underline{0.75}
    & \underline{1.53} & \underline{3.22} & \underline{0.86}
    & 5.39 & 7.71 & \underline{0.72} \\

    \rowcolor{oursrow}
    \textbf{PhysFlow (Ours)}
    & \textbf{4.05} & \textbf{7.32} & \textbf{0.76}
    & \textbf{1.38} & \textbf{3.12} & \textbf{0.87}
    & \textbf{5.23} & \textbf{7.36} & \textbf{0.74} \\

    \bottomrule
  \end{tabularx}
\end{table*}

\begin{figure}[t]   
  \centering
  \setlength{\fboxsep}{0pt} 
  \includegraphics[width=\linewidth]{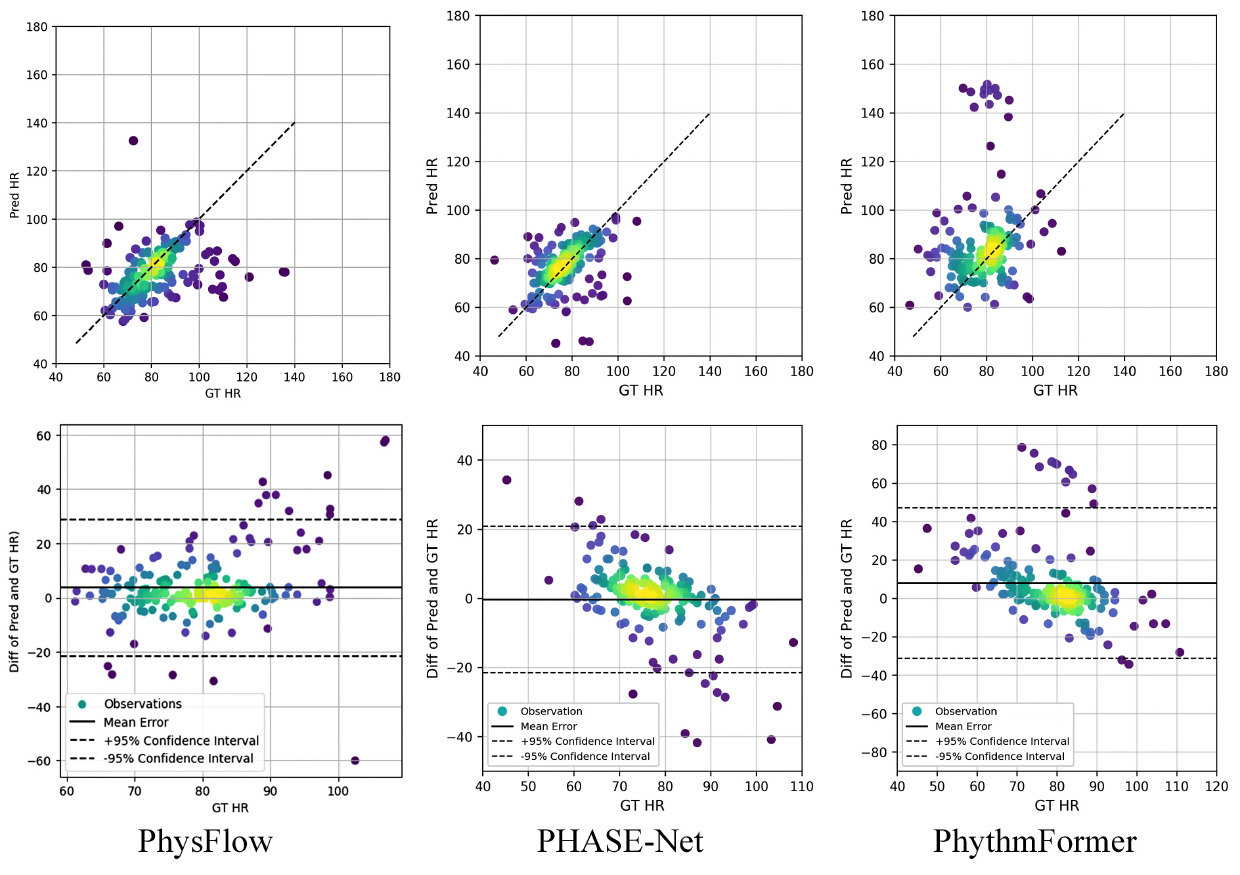} 
\caption{Qualitative comparison of HR estimation results on the BUAA-MIHR dataset. The top row shows scatter plots of predicted HR and ground-truth HR, and the bottom row shows the corresponding Bland--Altman plots for PhysFlow, PHASE-Net and RhythmFormer.}
  \label{fig:32}
\end{figure}

\begin{table}[t]
	\centering
	\caption{Waveform-level quantitative evaluation on the BUAA dataset.}
	\label{waveform_level_form}
	\small
	\renewcommand{\arraystretch}{1.08}
	\setlength{\tabcolsep}{5pt}
	\begin{tabular*}{\linewidth}{@{\extracolsep{\fill}}lccc@{}}
		\toprule
		Method & Wave-Corr$\uparrow$ & PE$\downarrow$ & SNR$\uparrow$ \\
		\midrule
		RhythmMamba & 0.84 & 0.71 & 12.05 \\
		RhythmFormer & 0.85 & 0.69 & 11.64 \\
		PHASE-Net    & 0.89 & 0.37 & 12.43 \\
		\textbf{PhysFlow} & \textbf{0.95} & \textbf{0.12} & \textbf{17.18} \\
		\bottomrule
	\end{tabular*}
\end{table}
\subsection{Intra-dataset Evaluation}
The quantitative results of intra-dataset evaluation are summarized in Table.~\ref{tab:intra_dataset}. Overall, PhysFlow attains the best performance on all four benchmarks, including NIRP-IND, NIRP-DRV, VIPL-HR, and BUAA, which demonstrates its strong robustness across both relatively controlled and highly challenging recording conditions. Under the NIRP setting, where NIRP-IND and NIRP-DRV are jointly used for training, PhysFlow obtains the lowest estimation errors on both subsets. Specifically, it achieves RMSE values of 0.89 on NIRP-IND and 5.93 on NIRP-DRV, outperforming the strongest competing method PHASE-Net, which yields RMSE values of 0.98 and 6.23. These results indicate that PhysFlow can recover reliable physiological signals in both the relatively controlled indoor setting and the more challenging driving scenario. On the standard intra-dataset evaluations of VIPL-HR and BUAA, PhysFlow likewise delivers the best overall performance. On VIPL-HR, it achieves an RMSE of 5.79, improving over the second-best result of 6.37 obtained by PHASE-Net. On BUAA, PhysFlow further reduces the RMSE to 1.17, compared with the previous best result of 1.43. Similar gains can also be observed in MAE and Pearson's correlation coefficient. These improvements across datasets with different characteristics suggest that the proposed PhysFlow is effective in suppressing complex disturbances while preserving the physiological dynamics required for accurate rPPG estimation.

\begin{figure}[t]   
  \centering
  \setlength{\fboxsep}{0pt} 
  \includegraphics[width=\linewidth]{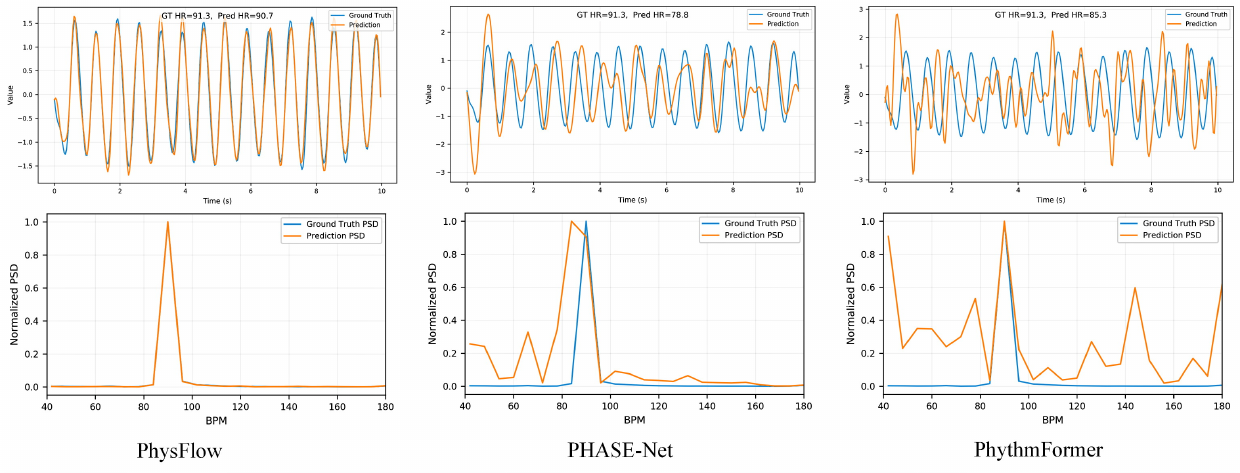} 
  \caption{Qualitative comparison of reconstructed rPPG waveforms on the BUAA-MIHR dataset. The top row shows the predicted and ground-truth rPPG signals in the time domain, and the bottom row shows their PSD distributions.}
  \label{fig:34}
\end{figure}

To further examine the behavior of different methods under intra-dataset evaluation, we provide both visual and waveform-level quantitative comparisons on the BUAA dataset. As shown in Fig.~\ref{fig:32}, the scatter plots of predicted HR versus ground-truth HR indicate that the estimates of PhysFlow are distributed more closely around the identity line than those of PHASE-Net and RhythmFormer, suggesting a stronger agreement with the reference HR values. The corresponding Bland--Altman plots further show that PhysFlow yields a smaller estimation bias and a narrower limit of agreement, indicating more accurate and more stable HR estimation across different HR ranges. Fig.~\ref{fig:34} further presents representative waveform-level comparisons together with their corresponding power spectral density (PSD) distributions. PhysFlow produces reconstructed rPPG waveforms that more closely match the ground truth in terms of periodic pattern, temporal alignment, and overall morphology. In the frequency domain, its dominant spectral peak is also more consistent with the ground-truth HR frequency, whereas the competing methods exhibit more evident spectral dispersion and spurious peaks. By contrast, PHASE-Net and RhythmFormer show more noticeable local distortions, irregular fluctuations, and phase deviations in the reconstructed waveforms. Overall, these qualitative results demonstrate that the advantage of PhysFlow lies not only in improved HR estimation accuracy, but also in more faithful waveform reconstruction.

\begin{figure*}[t]   
  \centering
  \setlength{\fboxsep}{0pt} 
  \includegraphics[width=\linewidth]{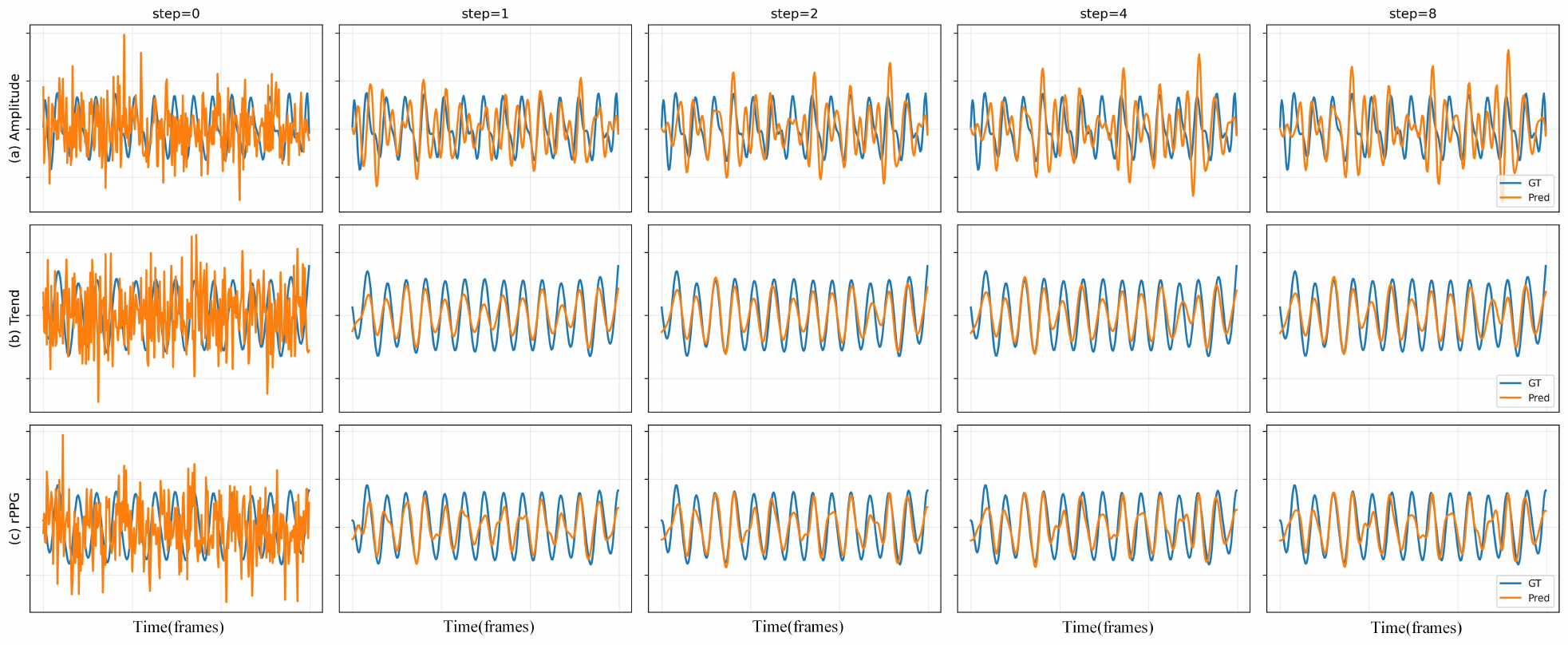} 
\caption{Visualization of the generated amplitude component, trend component and reconstructed rPPG signal under different inference steps on the NIRP-DRV dataset. From left to right, the columns correspond to step=0, 1, 2, 4, and 8. As the inference step increases, the generated signals become progressively closer to the ground truth.}
  \label{fig:33}
\end{figure*}

\subsection{Cross-dataset Evaluation}
The cross-dataset comparison results are summarized in Table.~\ref{tab:cross_dataset}. Compared with intra-dataset evaluation, all methods exhibit noticeable performance degradation under cross-dataset transfer, confirming the intrinsic difficulty of generalization across different acquisition devices, recording environments, and subject distributions. Nevertheless, PhysFlow maintains the best overall performance in all evaluated cross-dataset settings, indicating superior generalization capability. When trained on NIRP and tested on VIPL-HR, PhysFlow achieves the best results with an RMSE of 7.32 and a Pearson correlation of 0.76, slightly improving over the strongest competing methods. More importantly, under the reverse transfer setting from VIPL-HR to NIRP, PhysFlow outperforms all compared methods on both NIRP-IND and NIRP-DRV. It achieves RMSE values of 3.12 and 7.36, compared with the second-best results of 3.22 and 7.52. The improvements are also reflected in MAE and correlation coefficient. These results suggest that PhysFlow learns more stable and transferable physiological representations than existing methods. By explicitly decomposing the supervision signal into trend and amplitude components and modeling them with separate conditional velocity fields, PhysFlow reduces the sensitivity of waveform reconstruction to dataset-specific nuisance factors, thereby improving robustness under cross-dataset distribution shifts.

Beyond HR-level visualization, Table~\ref{waveform_level_form} reports waveform-level quantitative results from morphology, phase, and spectral perspectives. PhysFlow achieves the highest Wave-Corr of 0.95, the lowest PE of 0.12 and the highest SNR of 17.18, consistently outperforming RhythmMamba, RhythmFormer, and PHASE-Net across all three metrics. These results indicate that the reconstructed waveforms of PhysFlow are not only more morphologically consistent with the ground truth, but also exhibit better temporal phase alignment and cleaner spectral concentration around the physiological pulse frequency. Such observations are well aligned with the visual comparisons in Fig.~\ref{fig:34}, and further confirm the superiority of PhysFlow in preserving both the global waveform structure and the underlying pulse-related characteristics.

\begin{table}[htbp] 
  \centering
  \caption{Computational efficiency comparison. The best results are highlighted in \textbf{bold}, and the second-best results are underlined.}
  \label{tab:efficiency}
  \resizebox{\linewidth}{!}{%
    \begin{tabular}{lccccc}
      \toprule
      \textbf{Method} & 
      \textbf{Params (M)}$\downarrow$ & 
      \textbf{Flops (G)}$\downarrow$ & 
      \textbf{Throughput (Kps)}$\uparrow$ & 
      \textbf{Time (ms)}$\downarrow$ & 
      \textbf{Mem (M)}$\downarrow$ \\
      \midrule
      DeepPhys\cite{DeepPhys}  & \underline{1.98} & 111.67 & 28.89 & 34.61 & 10638 \\
      TS-CAN \cite{TS-CAN}      & 3.91 & 110.15 & 26.23 & 38.13 & 11834 \\
        ND-DeeprPPG \cite{16}        & 6.05 & 320.08 & 22.45 & 43.36 & 14327 \\

      PhysFormer++ \cite{Physformer++}     & 7.38 & 47.44  & 50.79 & 19.69 & 6480 \\
      EfficientPhys \cite{Efficientphys} & \textbf{1.91} & 56.06  & 41.36 & 24.18 & 7814 \\
      RhythmFormer \cite{Rhythmformer} & 3.33 & 76.99 & 11.10 & \underline{14.41} &\underline{726} \\
      RhythmMamba \cite{Rhythmmamba}  & 2.00 & \textbf{12.41} & 27.16 & 36.82 & 2450 \\
      PhysDiff \cite{17}      & 2.64 & \underline{22.46}  & \underline{60.23} & 16.60 & 1246 \\
      PHASE-Net \cite{43}      & 3.31 & 72.65  & 56.19& 26.41 & 2131 \\
      \midrule
      \textbf{PhysFlow (Ours)}         & 3.86 & 27.75 & \textbf{63.74} & \textbf{13.63} & \textbf{614} \\
      \bottomrule
    \end{tabular}%
  }
\end{table}

To evaluate the efficiency of PhysFlow, we conduct a 10-second inference test on a single NVIDIA RTX 4090 GPU, and summarize the results in Table.~\ref{tab:efficiency}. Although PhysFlow does not have the smallest parameter count or FLOPs, it achieves the highest throughput, the lowest inference time, and the lowest memory consumption. These results demonstrate that PhysFlow maintains strong estimation performance while offering favorable computational efficiency for real-time rPPG applications.

\begin{table}[htbp]
	\caption{HR estimation results of the inference-step ablation experiment conducted on the NIRP-DRV dataset.}
	\label{tab:step_ablation}
	\centering
	\begin{tabular*}{0.5\textwidth}{@{\extracolsep{\fill}}cccc}
		\toprule
		Step & MAE $\downarrow$ & RMSE $\downarrow$ & $\rho \uparrow$ \\
		\midrule
		1 & $3.27\pm0.47$ & $6.42\pm0.46$ & $0.84\pm0.13$ \\
		2 & $3.13\pm0.34$ & $6.18\pm0.39$ & $0.85\pm0.09$ \\
		4 & $2.63\pm0.14$ & $5.93\pm0.12$ & $0.87\pm0.02$ \\
		8 & $2.59\pm0.12$ & $5.89\pm0.10$ & $0.87\pm0.02$ \\
		\bottomrule
	\end{tabular*}
\end{table}

\begin{table}[t]
  \caption{HR estimation results of the ablation study on the VIPL-HR dataset. PT denotes the periodic transformer. T-Head and A-Head denote the trend head and amplitude head. U-Head denotes a unified temporal residual convolutional velocity head. The joint use of $\mathcal{L}_{\mathrm{tre}}$ and $\mathcal{L}_{\mathrm{amp}}$ denoted the proposed dual-field loss $\mathcal{L}_{\mathrm{RF}}(\theta)$. The best results are highlighted in \textbf{bold}.}
  \label{tab:ablation}
  \centering
  \scriptsize
  \setlength{\tabcolsep}{4.5pt}
  \begin{tabular}{@{}ccccccccc@{}}
    \toprule
    PT & T-Head & A-Head & U-Head & $\mathcal{L}_{\mathrm{tre}}$ & $\mathcal{L}_{\mathrm{amp}}$ & MAE$\downarrow$ & RMSE$\downarrow$ & $\rho\uparrow$ \\
    \midrule
    \cmark & \cmark & \xmark & \xmark & \cmark & \xmark & 4.61 & 6.91 & 0.81 \\
    \cmark & \xmark & \cmark & \xmark & \xmark & \cmark & 4.56 & 6.83 & 0.82 \\
    \cmark & \cmark & \cmark & \xmark & \xmark & \cmark & 4.11 & 6.23 & 0.85 \\
    \cmark & \cmark & \cmark & \xmark & \cmark & \xmark & 3.92 & 5.94 & 0.85 \\
    \xmark & \cmark & \cmark & \xmark & \cmark & \cmark & 4.35 & 6.79 & 0.84 \\
    \cmark & \xmark & \xmark & \cmark & \xmark & \xmark & 4.73 & 7.13 & 0.80 \\
    \rowcolor{oursrow}

    \cmark & \cmark & \cmark & \xmark & \cmark & \cmark & \textbf{3.64} & \textbf{5.79} & \textbf{0.87} \\
    \bottomrule
  \end{tabular}
\end{table}

\subsection{Ablation studies}
We conduct a series of ablation studies to examine the contributions of different components in our method. We first examine the effect of the number of inference steps on HR prediction. As shown in Fig.~\ref{fig:33}, as the number of inference steps increases, the generated components and the reconstructed rPPG signal show better agreement with their corresponding targets. This suggests that additional rectified-flow integration steps improve the accuracy of signal generation and reconstruction. As shown in Table.~\ref{tab:step_ablation}, the performance improves as the number of inference steps increases and becomes nearly saturated at 4 steps on the NIRP-DRV dataset. Based on this observation, we fix the number of inference steps to 4 in the subsequent ablation experiments.

\begin{table}[t]
\centering
\caption{Additional analyses on PhysFlow.}
\label{tab:additional_analysis}
\renewcommand{\arraystretch}{1.03}
\setlength{\tabcolsep}{6pt}
\footnotesize
\begin{tabular}{lccc}
\toprule
Setting & MAE$\downarrow$ & RMSE$\downarrow$ & $\rho\uparrow$ \\
\midrule
\multicolumn{4}{l}{\textit{Decomposition study on VIPL-HR}} \\
Three-band split~\cite{47} & 5.01 & 8.41 & 0.81 \\
EMD~\cite{48}              & 4.67 & 7.94 & 0.82 \\
Wavelet~\cite{49}          & 4.71 & 8.13 & 0.82 \\
\rowcolor{oursrow}
Ours                        & \textbf{3.64} & \textbf{5.79} & \textbf{0.87} \\
\midrule
\multicolumn{4}{l}{\textit{Stability study on NIRP-IND}} \\
Random Gaussian             & 0.44 $\pm$ 0.08 & 0.90 $\pm$ 0.21 & 0.92 $\pm$ 0.03 \\
Canonical zero              & 0.46 & 0.94 & 0.91 \\
\bottomrule
\end{tabular}
\end{table}

\begin{table}[t]
	\caption{Sensitivity analysis of the key hyperparameters in FDD on the VIPL-HR dataset. For each parameter group, only one hyperparameter is varied while the others are fixed to their default settings. The best results in each group are highlighted in \textbf{bold}.}
	\label{tab:fdd_sensitivity}
	\centering
	\small
	\renewcommand{\arraystretch}{1.05}
	\setlength{\tabcolsep}{12pt}
	\begin{tabular}{ccccc}
		\toprule
		Parameter & Value & MAE$\downarrow$ & RMSE$\downarrow$ & $\rho\uparrow$ \\
		\midrule
		\multirow{3}{*}{$f_{tr}$}
		& 0.3  & 3.81 & 6.02 & 0.86 \\
		& \textbf{0.5}  & \textbf{3.64} & \textbf{5.79} & \textbf{0.87} \\
		& 0.7  & 3.76 & 5.95 & 0.86 \\
		\midrule
		\multirow{3}{*}{$f_{am}$}
		& 0.5  & 3.72 & 5.91 & 0.86 \\
		& \textbf{0.7}  & \textbf{3.64} & \textbf{5.79} & \textbf{0.87} \\
		& 0.9  & 3.79 & 6.01 & 0.86 \\
		\midrule
		\multirow{3}{*}{$\tau$}
		& 0.05 & 3.79 & 5.88 & 0.86 \\
		& \textbf{0.10} & \textbf{3.64} & \textbf{5.79} & \textbf{0.87} \\
		& 0.20 & 3.73 & 5.93 & 0.86 \\
        \midrule
        \multirow{3}{*}{$f_{\max}$}
        & 2.5  & 3.82 & 5.93 & 0.85 \\
        & \textbf{3.0}  & \textbf{3.64} & \textbf{5.79} & \textbf{0.87} \\
        & 3.5  & 3.73 & 5.87 & 0.84 \\

		\bottomrule
	\end{tabular}
\end{table}

Specifically, we compare different combinations of the trend head and amplitude head, together with their corresponding supervision losses, and further examine the contribution of the periodic transformer. For a fair comparison, when the periodic transformer is disabled, it is replaced with a vanilla ViT~\cite{41} while keeping the remaining framework unchanged. Similarly, when the proposed dual-field rectified flow loss is not used, the corresponding branch is optimized with the MSE loss. We also introduce a U-Head  by replacing the two component-specific velocity heads with a unified temporal residual convolutional velocity head, which models the concatenated trend and amplitude states and is trained with a single velocity-matching MSE objective against the concatenated target velocity. As shown in Table.~\ref{tab:ablation}, using only one head type leads to inferior performance. Even when both heads are enabled, applying only $\mathcal{L}_{\mathrm{tre}}$ or only $\mathcal{L}_{\mathrm{amp}}$ still fails to achieve the best results. The U-Head also performs worse than the full model, indicating that a shared unified velocity field is less effective than the proposed component-specific dual-field design. In contrast, jointly applying the two losses yields the best performance, demonstrating the effectiveness of the proposed rectified-flow loss $\mathcal{L}_{\mathrm{RF}}(\theta)$. Moreover, incorporating the periodic transformer further improves the overall performance. Overall, the best results are achieved by the full configuration that combines the trend head, amplitude head, periodic transformer and the loss $\mathcal{L}_{\mathrm{RF}}(\theta)$.

To further validate the proposed design, we conduct additional analyses on both the supervisory decomposition and the inference stability, as summarized in Table~\ref{tab:additional_analysis}. For the decomposition study, the proposed FDD outperforms three-band split, EMD decomposition, and wavelet decomposition under the same backbone, rectified-flow framework, and training settings, indicating that it provides more suitable supervisory targets for the two conditional velocity fields. For the stability analysis, we evaluate PhysFlow with multiple random Gaussian initializations and additionally include canonical zero initialization as a deterministic reference. For the random Gaussian setting, we report the mean and standard deviation over different evaluation seeds. The results show that PhysFlow exhibits limited performance variation across random seeds, while canonical zero initialization achieves comparable performance. This suggests that the reconstructed waveform is primarily determined by the video-conditioned velocity fields, rather than by the particular choice of the initial latent state.

To assess the effect of the proposed frequency-domain decomposition (FDD), we perform a sensitivity analysis on its key hyperparameters, including the trend cutoff $f_{tr}$, the amplitude lower cutoff $f_{am}$, the maximum frequency $f_{\max}$, and the softness coefficient $\tau$. As shown in Table~\ref{tab:fdd_sensitivity}, the performance varies with these hyperparameters, indicating that the decomposition is not entirely insensitive to parameter choices. Nevertheless, the variations remain within a reasonable range, and the default settings provide the best overall performance among the tested values. In particular, $f_{tr}=0.5$, $f_{am}=0.7$, and $\tau=0.10$ yield the lowest estimation errors and the highest correlation coefficient in their respective groups. A similar trend is observed for $f_{\max}$, where the default setting $f_{\max}=3.0$ achieves the best overall results. These findings suggest that the proposed FDD remains effective under moderate parameter variations, while the default settings yield the best overall performance.

\section{Conclusion}
\label{sec:conclusion}

In this paper, we propose PhysFlow, a frequency-decoupled dual-field rectified-flow framework for robust rPPG estimation. PhysFlow decomposes the ground-truth waveform into amplitude and trend components as supervisory targets, and learns separate conditional velocity fields to generate the corresponding outputs from Gaussian-initialized latent states, thereby reducing mutual interference under complex disturbances. This design improves both HR estimation accuracy and waveform reconstruction quality, especially in preserving overall waveform shape and subtle pulse morphology. Moreover, the rectified-flow formulation enables efficient inference with only a few ODE integration steps. Experiments show that PhysFlow outperforms existing state-of-the-art methods in estimation accuracy and waveform fidelity. Despite its strong performance, PhysFlow still has room for improvement in model complexity, which may limit its applicability to lightweight deployment scenarios. In addition, the frequency-domain decomposition strategy adopts fixed frequency cutoffs, which may limit its flexibility under abnormal heart rates or strong motion interference. In future work, we will focus on developing a lighter yet more effective feature extraction framework and adaptive decomposition strategy to further improve practical deployability and interpretability.

\section*{Acknowledgments}
This work was supported by the key project of Jiangsu Provincial Natural Science Fund under Grant No. BK20253028 and the National Nature Science Fund of China under Grant Nos. 62176124, U24A20330 and 62361166670.

\bibliographystyle{IEEEtran}
\bibliography{refernce}

@String(ECCV= {Eur. Conf. Comput. Vis.})

@String(ICME = {Int. Conf. Multimedia and Expo})

@String(AAAI = {AAAI})

@String(ECCV  = {ECCV})

@String(ICME  =	{ICME})

@String{Computer = "{IEEE} Computer" }

@String{Springer = "Springer-Verlag" }

@inproceedings{1,
  title={Remote Photoplethysmography in Real-World and Extreme Lighting Scenarios},
  author={Shao, Hang and Luo, Lei and Qian, Jianjun and Yan, Mengkai and Chen, Shuo and Yang, Jian},
  booktitle={Proceedings of the Computer Vision and Pattern Recognition Conference},
  pages={10858--10867},
  year={2025}
}

@inproceedings{2,
  title={Continual Learning for Remote Physiological Measurement: Minimize Forgetting and Simplify Inference},
  author={Liang, Qian and Chen, Yan and Hu, Yang},
  booktitle={European conference on computer vision},
  pages={126--144},
  year={2024},
  organization={Springer}
}

@article{3,
  title={A novel algorithm for remote photoplethysmography: Spatial subspace rotation},
  author={Wang, Wenjin and Stuijk, Sander and De Haan, Gerard},
  journal={IEEE transactions on biomedical engineering},
  volume={63},
  number={9},
  pages={1974--1984},
  year={2015},
  publisher={IEEE}
}

@article{4,
  title={PulseGAN: Learning to generate realistic pulse waveforms in remote photoplethysmography},
  author={Song, Rencheng and Chen, Huan and Cheng, Juan and Li, Chang and Liu, Yu and Chen, Xun},
  journal={IEEE Journal of Biomedical and Health Informatics},
  volume={25},
  number={5},
  pages={1373--1384},
  year={2021},
  publisher={IEEE}
}

@article{5,
  title={Unsupervised skin tissue segmentation for remote photoplethysmography},
  author={Bobbia, Serge and Macwan, Richard and Benezeth, Yannick and Mansouri, Alamin and Dubois, Julien},
  journal={Pattern recognition letters},
  volume={124},
  pages={82--90},
  year={2019},
  publisher={Elsevier}
}

@article{6,
  title={Noninvasive blood glucose monitoring using spatiotemporal ECG and PPG feature fusion and weight-based choquet integral multimodel approach},
  author={Li, Jingzhen and Ma, Jingjing and Omisore, Olatunji Mumini and Liu, Yuhang and Tang, Huajie and Ao, Pengfei and Yan, Yan and Wang, Lei and Nie, Zedong},
  journal={IEEE transactions on neural networks and learning systems},
  volume={35},
  number={10},
  pages={14491--14505},
  year={2023},
  publisher={IEEE}
}

@article{7,
  title={Transformer Meets Gated Residual Networks to Enhance PICU’s PPG Artifact Detection Informed by Mutual Information Neural Estimation},
  author={Le, Thanh-Dung and Macabiau, Clara and Albert, Kevin and Chatzinotas, Symeon and Jouvet, Philippe and Noumeir, Rita},
  journal={IEEE Transactions on Neural Networks and Learning Systems},
  year={2026},
  publisher={IEEE}
}

@inproceedings{8,
  title={3D mask face anti-spoofing with remote photoplethysmography},
  author={Liu, Siqi and Yuen, Pong C and Zhang, Shengping and Zhao, Guoying},
  booktitle={European Conference on Computer Vision},
  pages={85--100},
  year={2016},
  organization={Springer}
}

@inproceedings{9,
  title={Lstc-rppg: Long short-term convolutional network for remote photoplethysmography},
  author={Lee, Jun Seong and Hwang, Gyutae and Ryu, Moonwook and Lee, Sang Jun},
  booktitle={Proceedings of the IEEE/CVF Conference on Computer Vision and Pattern Recognition},
  pages={6015--6023},
  year={2023}
}

@inproceedings{10,
  title={Learning motion-robust remote photoplethysmography through arbitrary resolution videos},
  author={Li, Jianwei and Yu, Zitong and Shi, Jingang},
  booktitle={Proceedings of the AAAI Conference on Artificial Intelligence},
  volume={37},
  number={1},
  pages={1334--1342},
  year={2023}
}

@inproceedings{11,
  title={Automatic region-based heart rate measurement using remote photoplethysmography},
  author={Kossack, Benjamin and Wisotzky, Eric and Hilsmann, Anna and Eisert, Peter},
  booktitle={Proceedings of the IEEE/CVF International Conference on Computer Vision},
  pages={2755--2759},
  year={2021}
}

@article{12,
  title={A heart rate monitoring framework for real-world drivers using remote photoplethysmography},
  author={Huang, Po-Wei and Wu, Bing-Jhang and Wu, Bing-Fei},
  journal={IEEE journal of biomedical and health informatics},
  volume={25},
  number={5},
  pages={1397--1408},
  year={2020},
  publisher={IEEE}
}

@article{13,
  title={Robust remote photoplethysmography estimation with environmental noise disentanglement},
  author={Liu, Si-Qi and Yuen, Pong C},
  journal={IEEE Transactions on Image Processing},
  volume={33},
  pages={27--41},
  year={2023},
  publisher={IEEE}
}

@inproceedings{14,
  title={A novel framework for remote photoplethysmography pulse extraction on compressed videos},
  author={Zhao, Changchen and Lin, Chun-Liang and Chen, Weihai and Li, Zhengguo},
  booktitle={Proceedings of the IEEE conference on computer vision and pattern recognition workshops},
  pages={1299--1308},
  year={2018}
}

@article{15,
  title={Dual-path tokenlearner for remote photoplethysmography-based physiological measurement with facial videos},
  author={Qian, Wei and Guo, Dan and Li, Kun and Zhang, Xiaowei and Tian, Xilan and Yang, Xun and Wang, Meng},
  journal={IEEE Transactions on Computational Social Systems},
  volume={11},
  number={3},
  pages={4465--4477},
  year={2024},
  publisher={IEEE}
}

@inproceedings{16,
  title={Efficient remote photoplethysmography with temporal derivative modules and time-shift invariant loss},
  author={Comas, Joaquim and Ruiz, Adria and Sukno, Federico},
  booktitle={Proceedings of the IEEE/CVF conference on computer vision and pattern recognition},
  pages={2182--2191},
  year={2022}
}

@inproceedings{17,
  title={Physdiff: physiology-based dynamicity disentangled diffusion model for remote physiological measurement},
  author={Qian, Wei and Su, Gaoji and Guo, Dan and Zhou, Jinxing and Li, Xiaobai and Hu, Bin and Tang, Shengeng and Wang, Meng},
  booktitle={Proceedings of the AAAI Conference on Artificial Intelligence},
  volume={39},
  number={6},
  pages={6568--6576},
  year={2025}
}

@article{18,
  title={Spiking-PhysFormer: Camera-based remote photoplethysmography with parallel spike-driven transformer},
  author={Liu, Mingxuan and Tang, Jiankai and Chen, Yongli and Li, Haoxiang and Qi, Jiahao and Li, Siwei and Wang, Kegang and Gan, Jie and Wang, Yuntao and Chen, Hong},
  journal={Neural Networks},
  volume={185},
  pages={107128},
  year={2025},
  publisher={Elsevier}
}

@article{19,
  title={Tranphys: Spatiotemporal masked transformer steered remote photoplethysmography estimation},
  author={Shao, Hang and Luo, Lei and Qian, Jianjun and Chen, Shuo and Hu, Chuanfei and Yang, Jian},
  journal={IEEE Transactions on Circuits and Systems for Video Technology},
  volume={34},
  number={4},
  pages={3030--3042},
  year={2023},
  publisher={IEEE}
}

@inproceedings{20,
  title={Deep learning-based image enhancement for robust remote photoplethysmography in various illumination scenarios},
  author={Chen, Shutao and Ho, Sui Kei and Chin, Jing Wei and Luo, Kin Ho and Chan, Tsz Tai and So, Richard HY and Wong, Kwan Long},
  booktitle={Proceedings of the ieee/cvf conference on computer vision and pattern recognition},
  pages={6077--6085},
  year={2023}
}

@inproceedings{21,
  title={Analyzing Participants' Engagement during Online Meetings Using Unsupervised Remote Photoplethysmography with Behavioral Features},
  author={Vedernikov, Alexander and Sun, Zhaodong and Kykyri, Virpi-Liisa and Pohjola, Mikko and Nokia, Miriam and Li, Xiaobai},
  booktitle={Proceedings of the IEEE/CVF Conference on Computer Vision and Pattern Recognition},
  pages={389--399},
  year={2024}
}

@article{22,
  title={Tranpulse: Remote photoplethysmography estimation with time-varying supervision to disentangle multiphysiologically interference},
  author={Shao, Hang and Luo, Lei and Qian, Jianjun and Chen, Shuo and Hu, Chuanfei and Yang, Jian},
  journal={IEEE Transactions on Instrumentation and Measurement},
  volume={73},
  pages={1--11},
  year={2024},
  publisher={IEEE}
}

@article{23,
  title={A compensation network with error mapping for robust remote photoplethysmography in noise-heavy conditions},
  author={Wu, Bing-Fei and Wu, Yi-Chiao and Chou, Yi-Wei},
  journal={IEEE Transactions on Instrumentation and Measurement},
  volume={71},
  pages={1--11},
  year={2022},
  publisher={IEEE}
}

@article{24,
  title={Robust and remote photoplethysmography based on smartphone imaging of the human palm},
  author={Lian, Chao and Yang, Yiming and Yu, Xiaodong and Sun, Hui and Zhao, Yuliang and Zhang, Guanglie and Li, Wen Jung},
  journal={IEEE Transactions on Instrumentation and Measurement},
  volume={72},
  pages={1--11},
  year={2023},
  publisher={IEEE}
}

@article{rppg-toolbox,
  title={rppg-toolbox: Deep remote ppg toolbox},
  author={Liu, Xin and Narayanswamy, Girish and Paruchuri, Akshay and Zhang, Xiaoyu and Tang, Jiankai and Zhang, Yuzhe and Sengupta, Roni and Patel, Shwetak and Wang, Yuntao and McDuff, Daniel},
  journal={Advances in Neural Information Processing Systems},
  volume={36},
  pages={68485--68510},
  year={2023}
}

@inproceedings{26,
  title={Toward motion robustness: a masked attention regularization framework in remote photoplethysmography},
  author={Zhao, Pengfei and Sun, Qigong and Tian, Xiaolin and Yang, Yige and Tao, Shuo and Cheng, Jie and Chen, Jiantong},
  booktitle={Proceedings of the IEEE/CVF Conference on Computer Vision and Pattern Recognition},
  pages={7829--7838},
  year={2024}
}

@article{27,
  title={Realistic pulse waveforms estimation via contrastive learning in remote photoplethysmography},
  author={Dong, Bowei and Liu, Yuliang and Yang, Kaifeng and Cao, Jiajian},
  journal={IEEE Transactions on Instrumentation and Measurement},
  volume={73},
  pages={1--15},
  year={2024},
  publisher={IEEE}
}

@inproceedings{28,
  title={A LSTM-based realtime signal quality assessment for photoplethysmogram and remote photoplethysmogram},
  author={Gao, Haoyuan and Wu, Xiaopei and Shi, Chenyun and Gao, Qing and Geng, Jidong},
  booktitle={Proceedings of the IEEE/CVF Conference on Computer Vision and Pattern Recognition},
  pages={3831--3840},
  year={2021}
}

@article{29,
  title={Motion-resistant remote imaging photoplethysmography based on the optical properties of skin},
  author={Feng, Litong and Po, Lai-Man and Xu, Xuyuan and Li, Yuming and Ma, Ruiyi},
  journal={IEEE Transactions on Circuits and Systems for Video Technology},
  volume={25},
  number={5},
  pages={879--891},
  year={2014},
  publisher={IEEE}
}

@article{30,
  title={Deep-learning-based remote photoplethysmography measurement in driving scenarios with color and near-infrared images},
  author={Chiu, Li-Wen and Chou, Yang-Ren and Wu, Yi-Chiao and Wu, Bing-Fei},
  journal={IEEE Transactions on Instrumentation and Measurement},
  volume={72},
  pages={1--12},
  year={2023},
  publisher={IEEE}
}

@inproceedings{31,
  title={PhysFFTFormer: A Frequency Domain-based Vision Transformer for Efficient Remote Physiological Measurement},
  author={Liu, Fangyuan and Zhao, Sirui and Xu, Tong and Sun, Yu and Wang, Hao and Zhang, Suojuan and Chen, Enhong},
  booktitle={2025 IEEE International Conference on Multimedia and Expo (ICME)},
  pages={1--6},
  year={2025},
  organization={IEEE}
}

@article{32,
  title={Flow straight and fast: Learning to generate and transfer data with rectified flow},
  author={Liu, Xingchao and Gong, Chengyue and Liu, Qiang},
  journal={arXiv preprint arXiv:2209.03003},
  year={2022}
}

@inproceedings{33,
  title={Flowgrad: Controlling the output of generative odes with gradients},
  author={Liu, Xingchao and Wu, Lemeng and Zhang, Shujian and Gong, Chengyue and Ping, Wei and Liu, Qiang},
  booktitle={Proceedings of the IEEE/CVF Conference on Computer Vision and Pattern Recognition},
  pages={24335--24344},
  year={2023}
}

@inproceedings{34,
  title={Flowie: Efficient image enhancement via rectified flow},
  author={Zhu, Yixuan and Zhao, Wenliang and Li, Ao and Tang, Yansong and Zhou, Jie and Lu, Jiwen},
  booktitle={Proceedings of the IEEE/CVF Conference on Computer Vision and Pattern Recognition},
  pages={13--22},
  year={2024}
}

@inproceedings{35,
  title={Instaflow: One step is enough for high-quality diffusion-based text-to-image generation},
  author={Liu, Xingchao and Zhang, Xiwen and Ma, Jianzhu and Peng, Jian and others},
  booktitle={The Twelfth International Conference on Learning Representations},
  year={2023}
}

@article{36,
  title={One diffusion step to real-world super-resolution via flow trajectory distillation},
  author={Li, Jianze and Cao, Jiezhang and Guo, Yong and Li, Wenbo and Zhang, Yulun},
  journal={arXiv preprint arXiv:2502.01993},
  year={2025}
}

@inproceedings{37,
  title={VIPL-HR: A multi-modal database for pulse estimation from less-constrained face video},
  author={Niu, Xuesong and Han, Hu and Shan, Shiguang and Chen, Xilin},
  booktitle={Asian conference on computer vision},
  pages={562--576},
  year={2018},
  organization={Springer}
}

@article{38,
  title={Rhythmnet: End-to-end heart rate estimation from face via spatial-temporal representation},
  author={Niu, Xuesong and Shan, Shiguang and Han, Hu and Chen, Xilin},
  journal={IEEE Transactions on Image Processing},
  volume={29},
  pages={2409--2423},
  year={2019},
  publisher={IEEE}
}

@inproceedings{39,
  title={Image enhancement for remote photoplethysmography in a low-light environment},
  author={Xi, Lin and Chen, Weihai and Zhao, Changchen and Wu, Xingming and Wang, Jianhua},
  booktitle={2020 15th IEEE International Conference on Automatic Face and Gesture Recognition (FG 2020)},
  pages={1--7},
  year={2020},
  organization={IEEE}
}

@inproceedings{40-indoor,
  title={SparsePPG: Towards driver monitoring using camera-based vital signs estimation in near-infrared},
  author={Magdalena Nowara, Ewa and Marks, Tim K and Mansour, Hassan and Veeraraghavan, Ashok},
  booktitle={Proceedings of the IEEE conference on computer vision and pattern recognition workshops},
  pages={1272--1281},
  year={2018}
}

@article{40-drv,
  title={Near-infrared imaging photoplethysmography during driving},
  author={Nowara, Ewa M and Marks, Tim K and Mansour, Hassan and Veeraraghavan, Ashok},
  journal={IEEE transactions on intelligent transportation systems},
  volume={23},
  number={4},
  pages={3589--3600},
  year={2020},
  publisher={IEEE}
}

@article{pos,
  title={Single-element remote-ppg},
  author={Wang, Wenjin and Den Brinker, Albertus C and De Haan, Gerard},
  journal={IEEE Transactions on Biomedical Engineering},
  volume={66},
  number={7},
  pages={2032--2043},
  year={2018},
  publisher={IEEE}
}

@article{chrom,
  title={Robust pulse rate from chrominance-based rPPG},
  author={De Haan, Gerard and Jeanne, Vincent},
  journal={IEEE transactions on biomedical engineering},
  volume={60},
  number={10},
  pages={2878--2886},
  year={2013},
  publisher={IEEE}
}

@inproceedings{LGI,
  title={Local group invariance for heart rate estimation from face videos in the wild},
  author={Pilz, Christian S and Zaunseder, Sebastian and Krajewski, Jarek and Blazek, Vladimir},
  booktitle={Proceedings of the IEEE conference on computer vision and pattern recognition workshops},
  pages={1254--1262},
  year={2018}
}

@inproceedings{DeepPhys,
  title={Deepphys: Video-based physiological measurement using convolutional attention networks},
  author={Chen, Weixuan and McDuff, Daniel},
  booktitle={Proceedings of the european conference on computer vision (ECCV)},
  pages={349--365},
  year={2018}
}

@article{TS-CAN,
  title={Multi-task temporal shift attention networks for on-device contactless vitals measurement},
  author={Liu, Xin and Fromm, Josh and Patel, Shwetak and McDuff, Daniel},
  journal={Advances in Neural Information Processing Systems},
  volume={33},
  pages={19400--19411},
  year={2020}
}

@inproceedings{PFE-TFA,
  title={Learning motion-robust remote photoplethysmography through arbitrary resolution videos},
  author={Li, Jianwei and Yu, Zitong and Shi, Jingang},
  booktitle={Proceedings of the AAAI Conference on Artificial Intelligence},
  volume={37},
  number={1},
  pages={1334--1342},
  year={2023}
}

@inproceedings{NEST,
  title={Neuron structure modeling for generalizable remote physiological measurement},
  author={Lu, Hao and Yu, Zitong and Niu, Xuesong and Chen, Ying-Cong},
  booktitle={Proceedings of the IEEE/CVF conference on computer vision and pattern recognition},
  pages={18589--18599},
  year={2023}
}

@inproceedings{Efficientphys,
  title={Efficientphys: Enabling simple, fast and accurate camera-based cardiac measurement},
  author={Liu, Xin and Hill, Brian and Jiang, Ziheng and Patel, Shwetak and McDuff, Daniel},
  booktitle={Proceedings of the IEEE/CVF winter conference on applications of computer vision},
  pages={5008--5017},
  year={2023}
}

@article{Physformer++,
  title={Physformer++: Facial video-based physiological measurement with slowfast temporal difference transformer},
  author={Yu, Zitong and Shen, Yuming and Shi, Jingang and Zhao, Hengshuang and Cui, Yawen and Zhang, Jiehua and Torr, Philip and Zhao, Guoying},
  journal={International Journal of Computer Vision},
  volume={131},
  number={6},
  pages={1307--1330},
  year={2023},
  publisher={Springer}
}

@article{Rhythmformer,
  title={Rhythmformer: Extracting patterned rppg signals based on periodic sparse attention},
  author={Zou, Bochao and Guo, Zizheng and Chen, Jiansheng and Zhuo, Junbao and Huang, Weiran and Ma, Huimin},
  journal={Pattern Recognition},
  volume={164},
  pages={111511},
  year={2025},
  publisher={Elsevier}
}

@inproceedings{Rhythmmamba,
  title={Rhythmmamba: Fast, lightweight, and accurate remote physiological measurement},
  author={Zou, Bochao and Guo, Zizheng and Hu, Xiaocheng and Ma, Huimin},
  booktitle={Proceedings of the AAAI Conference on Artificial Intelligence},
  volume={39},
  number={10},
  pages={11077--11085},
  year={2025}
}

@article{41,
  title={An image is worth 16x16 words: Transformers for image recognition at scale},
  author={Dosovitskiy, Alexey and Beyer, Lucas and Kolesnikov, Alexander and Weissenborn, Dirk and Zhai, Xiaohua and Unterthiner, Thomas and Dehghani, Mostafa and Minderer, Matthias and Heigold, Georg and Gelly, Sylvain and others},
  journal={arXiv preprint arXiv:2010.11929},
  year={2020}
}

@article{42,
  title={Physllm: Harnessing large language models for cross-modal remote physiological sensing},
  author={Xie, Yiping and Zhao, Bo and Dai, Mingtong and Zhou, Jian-Ping and Sun, Yue and Tan, Tao and Xie, Weicheng and Shen, Linlin and Yu, Zitong},
  journal={arXiv preprint arXiv:2505.03621},
  year={2025}
}

@inproceedings{43,
  title={Phase-net: Physics-grounded harmonic attention system for efficient remote photoplethysmography measurement},
  author={Zhao, Bo and Guo, Dan and Cao, Junzhe and Xu, Yong and Zou, Bochao and Tan, Tao and Sun, Yue and Yu, Zitong},
  booktitle={Proceedings of the IEEE/CVF Conference on Computer Vision and Pattern Recognition},
  pages={21198--21207},
  year={2026}
}

@inproceedings{44,
  title={Physformer: Facial video-based physiological measurement with temporal difference transformer},
  author={Yu, Zitong and Shen, Yuming and Shi, Jingang and Zhao, Hengshuang and Torr, Philip HS and Zhao, Guoying},
  booktitle={Proceedings of the IEEE/CVF conference on computer vision and pattern recognition},
  pages={4186--4196},
  year={2022}
}

@article{45,
  title={Video-based Instantaneous Heart Rate Measurement with Enhanced Time-Frequency Representations},
  author={Cheng, Juan and Luo, Xiwen and Wu, Xiaowei and Song, Rencheng and Liu, Yu},
  journal={IEEE Transactions on Multimedia},
  year={2025},
  publisher={IEEE}
}

@inproceedings{46,
  title={Dual-gan: Joint bvp and noise modeling for remote physiological measurement},
  author={Lu, Hao and Han, Hu and Zhou, S Kevin},
  booktitle={Proceedings of the IEEE/CVF conference on computer vision and pattern recognition},
  pages={12404--12413},
  year={2021}
}

@article{47,
  title={Robust heart rate from fitness videos},
  author={Wang, Wenjin and den Brinker, Albertus C and Stuijk, Sander and de Haan, Gerard},
  journal={Physiological measurement},
  volume={38},
  number={6},
  pages={1023--1044},
  year={2017},
  publisher={IOP Publishing}
}

@article{48,
  title={Remote measurement of heart rate from facial video in different scenarios},
  author={Zheng, Xiujuan and Zhang, Chang and Chen, Hui and Zhang, Yun and Yang, Xiaomei},
  journal={Measurement},
  volume={188},
  pages={110243},
  year={2022},
  publisher={Elsevier}
}

@article{49,
  title={Detail-preserving arterial pulse wave measurement based biorthogonal wavelet decomposition from remote rgb observations},
  author={Tong, Yonggang and Huang, Zhipei and Zhang, Zhen and Yin, Ming and Shan, Guangcun and Wu, Jiankang and Qin, Fei},
  journal={Measurement},
  volume={222},
  pages={113605},
  year={2023},
  publisher={Elsevier}
}

@article{50,
  title={Video Respiratory Rate Measurement in Walking Scenarios Using Multi-strategy Adaptive Denoising},
  author={Pei, Gan and Ning, Junhao and Niu, Chenrui and Yao, Siqiong and Hu, Menghan and Zhai, Guangtao},
  journal={IEEE Transactions on Circuits and Systems for Video Technology},
  year={2026},
  publisher={IEEE}
}

@article{52,
  title={Video-based multiphysiological disentanglement and remote robust estimation for respiration},
  author={Shao, Hang and Luo, Lei and Qian, Jianjun and Yan, Mengkai and Gao, Shangbing and Yang, Jian},
  journal={IEEE Transactions on Neural Networks and Learning Systems},
  volume={36},
  number={5},
  pages={8360--8371},
  year={2024},
  publisher={IEEE}
}

@article{53,
  title={SaTPhys: Sandglass Transformer for Efficient Video-based Remote Physiological Measurement},
  author={Chu, Shuyang and Shi, Jingang and Yuan, Mengyao and Li, Xuqi and Jiang, Zhengdong and Zhao, Guoying},
  journal={IEEE Transactions on Circuits and Systems for Video Technology},
  year={2026},
  publisher={IEEE}
}

@article{54,
  title={Neuron perception inspired EEG emotion recognition with parallel contrastive learning},
  author={Li, Dongdong and Huang, Shengyao and Xie, Li and Wang, Zhe and Xu, Jiazhen},
  journal={IEEE transactions on neural networks and learning systems},
  year={2025},
  publisher={IEEE}
}

@article{55,
  title={Amplitude--time dual-view fused EEG temporal feature learning for automatic sleep staging},
  author={An, Panfeng and Zhao, Jianhui and Du, Bo and Zhao, Wenyuan and Zhang, Tingbao and Yuan, Zhiyong},
  journal={IEEE Transactions on Neural Networks and Learning Systems},
  volume={35},
  number={5},
  pages={6492--6506},
  year={2022},
  publisher={IEEE}
}

@article{56,
  title={Adversarial spatiotemporal contrastive learning for electrocardiogram signals},
  author={Wang, Ning and Feng, Panpan and Ge, Zhaoyang and Zhou, Yanjie and Zhou, Bing and Wang, Zongmin},
  journal={IEEE transactions on neural networks and learning systems},
  volume={35},
  number={10},
  pages={13845--13859},
  year={2023},
  publisher={IEEE}
}

\begin{IEEEbiography}[{\includegraphics[width=1in,height=1.25in,clip,keepaspectratio]{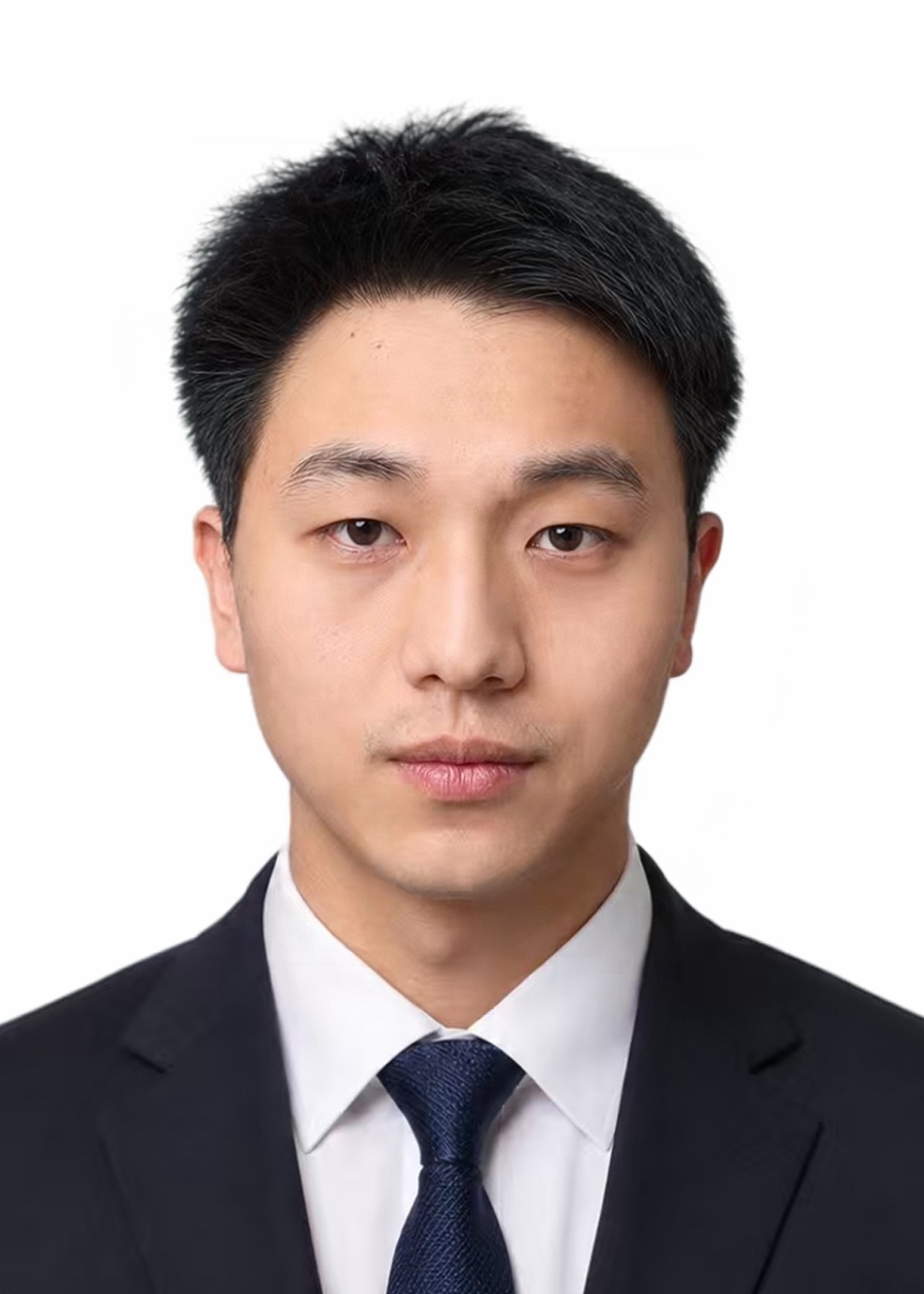}}]
{Zixu Li} received the B.S. degree in software engineering from the Business School of Jiang Xi University of Science and Technology, Nanchang, China, in 2022. He received the M.S. degree in electronic information with the School of Information and Electronic Engineering, Shandong Technology and Business University, Yantai, China, in 2025. He is currently pursuing the Ph.D. degree with the School of Computer Science and Engineering, Nanjing University of Science and Technology, Nanjing, China. 
His research interests include computer vision and pattern recognition, with a particular focus on physiological information measurement.
\end{IEEEbiography}

\begin{IEEEbiography}[{\includegraphics[width=1in,height=1.25in,clip,keepaspectratio]{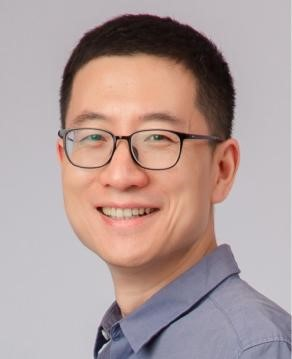}}]
{Jianjun Qian} (Member,IEEE) received the PhD degree from Nanjing University of Science and Technology (NUST), on the subject of pattern recognition and intelligence systems in 2014. He was selected as a Hong Kong Scholar in China in 2018. He is currently a Professor with the Key Lab oratory of Intelligent Perception and Systems for High-Dimensional Information of Ministry of Ed ucation, School of Computer Science and Engineer ing, NUST. His current research interests include pattern recognition, computer vision, and machine learning. He has served as Guest Editors for Neural Processing Letter and The Visual Computer.
\end{IEEEbiography}

\begin{IEEEbiography}[{\includegraphics[width=1in,height=1.25in,clip,keepaspectratio]{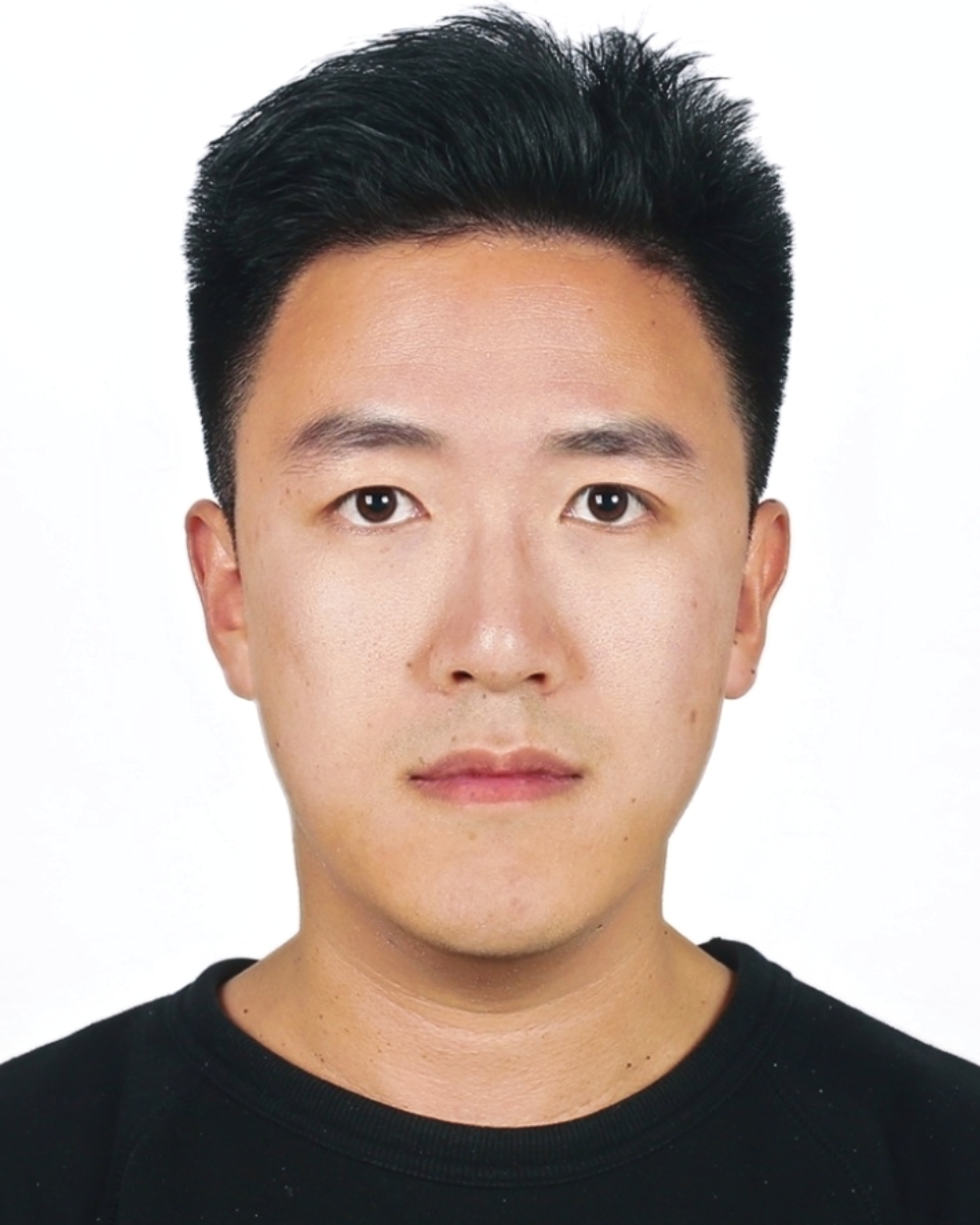}}]
{Hang Shao} received the M.S. and Ph.D. degrees from University of Shanghai for Science and Technology, Shanghai, China, and Nanjing University of Science and Technology, Nanjing, China, in 2021 and 2025, respectively, with a focus on pattern recognition and intelligent systems. He is currently a Lecturer with Qingdao University, Qingdao, China. His research interests include computer vision and biometric recognition. He has served as a reviewer for IEEE TIP, TNNLS, TCSVT, TFS and TMM.	
\end{IEEEbiography}

\begin{IEEEbiography}[{\includegraphics[width=1in,height=1.25in,clip,keepaspectratio]{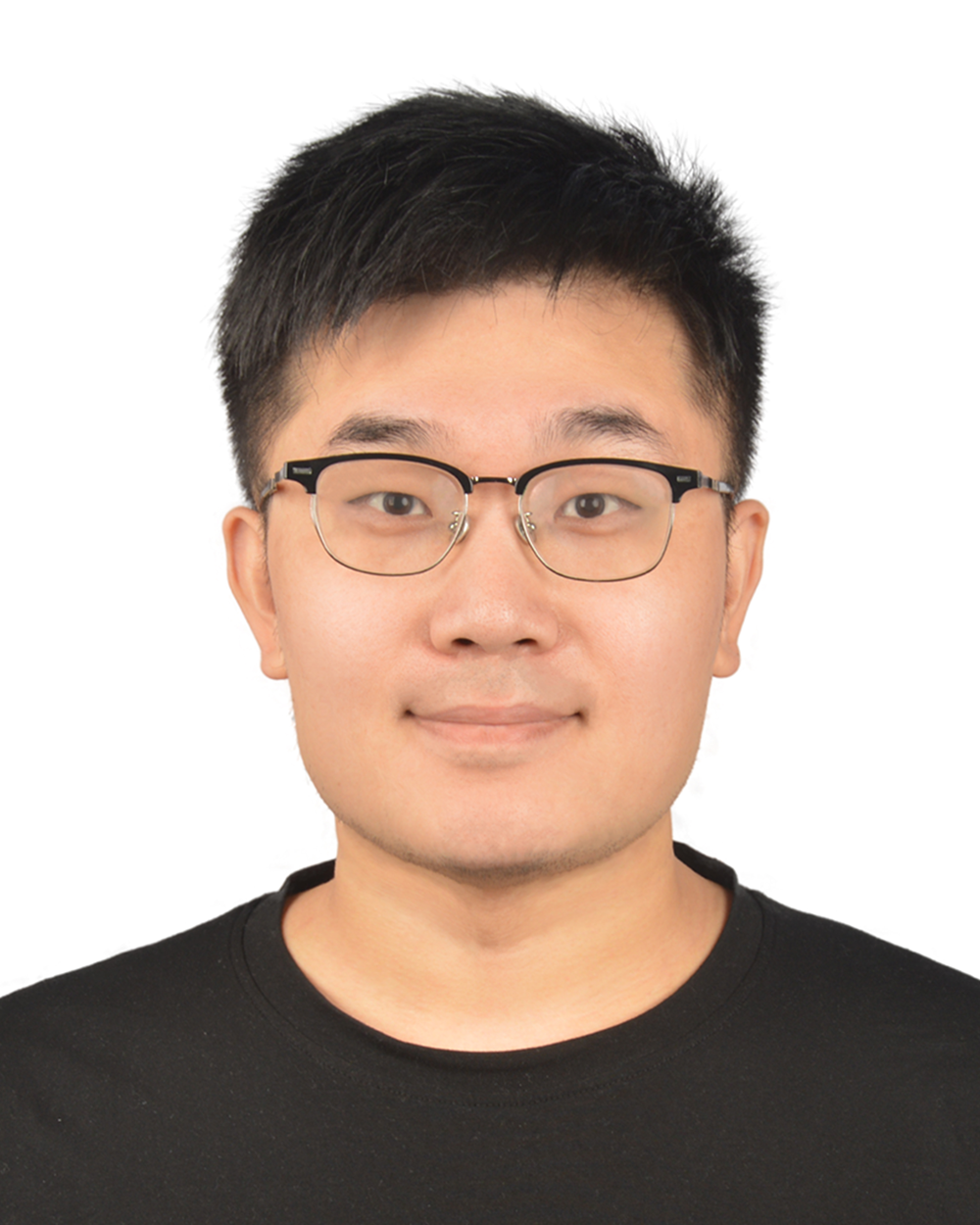}}]
{Lei Luo} received the Ph.D. degree in pattern recognition and intelligence systems from Nanjing University of Science and Technology (NJUST), Nanjing, China, in 2016. From 2017 to 2020, he was a Post-Doctoral Fellow at University of Texas at Arlington, TX, USA, and University of Pittsburgh, PA, USA. He is currently a Professor with NJUST. His research interests include pattern recognition, machine learning, data mining, and computer vision. He has served as an PC/SPC Member for CVPR, ICML, AAAI, IJCAI, NeurIPS, KDD, and ECCV, and a reviewer for multiple international journals, such as IEEE TPAMI, TIP, TSP, TNNLS, TKDE, TCSVT, and TMM.

\end{IEEEbiography}

\begin{IEEEbiography}[{\includegraphics[width=1in,height=1.25in,clip,keepaspectratio]{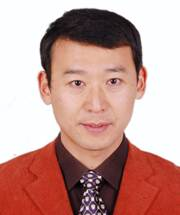}}]
{Jian Yang} received the PhD degree from Nanjing University of Science and Technology (NUST) in 2002, majoring in pattern recognition and intelligence systems. From 2003 to 2007, he was a Postdoctoral Fellow at the University of Zaragoza, Hong Kong Polytechnic University and New Jersey Institute of Technology, respectively. From 2007 to present, he is a professor in the School of Computer Science and Technology of NUST. His papers have been cited over 50000 times in the Scholar Google. His research interests include pattern recognition and computer vision. Currently, he is/was an associate editor of Pattern Recognition, Pattern Recognition Letters, IEEE Trans. Neural Networks and Learning Systems, and Neurocomputing. He is a Fellow of IAPR. 
\end{IEEEbiography}

\vfill
\clearpage

\end{document}